\documentclass[runningheads]{llncs}

\usepackage{eccv}

\usepackage{eccvabbrv}

\usepackage{graphicx}
\usepackage{booktabs}
\usepackage{xcolor}
\usepackage{booktabs}
\usepackage{colortbl}
\usepackage{bbding}
\usepackage{multirow}
\usepackage{multicol}
\usepackage{hyperref}
\usepackage{enumitem}
\usepackage[table]{xcolor}
\usepackage{wrapfig}
\usepackage{makecell}
\usepackage{pifont}
\usepackage{lipsum}
\usepackage{pifont}
\usepackage{multirow}
\usepackage{titletoc}

\usepackage[accsupp]{axessibility}  
\newcommand{\proposed}{\textsf{SToP}}
\definecolor{gainsboro}{RGB}{233,233,233}

\usepackage{hyperref}

\usepackage{orcidlink}

\begin{document}

\title{Sink-Token-Aware Pruning for Fine-Grained Video Understanding in Efficient Video LLMs} 

\titlerunning{SToP: Sink-Token-Aware Pruning}




\author{Kibum Kim$^{1, \href{mailto:kb.kibum@kaist.ac.kr}{\includegraphics[width=0.23cm]{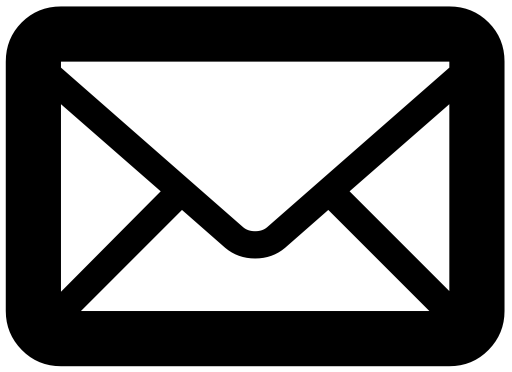}}}$ \and
Jiwan Kim$^{1, \href{mailto:kim.jiwan@kaist.ac.kr}{\includegraphics[width=0.23cm]{Figure/mail.png}}}$ \and
Kyle Min$^{2, \href{mailto:kyle.min@oracle.com}{\includegraphics[width=0.23cm]{Figure/mail.png}}}$ \and
Yueqi Wang$^{3, \href{mailto:yueqi@berkeley.edu}{\includegraphics[width=0.23cm]{Figure/mail.png}}}$ \\ \and
Jinyoung Moon$^{4, \href{mailto:jymoon@etri.re.kr}{\includegraphics[width=0.23cm]{Figure/mail.png}}}$ \and
Julian McAuley$^{3, \href{mailto:jmcauley@ucsd.edu}{\includegraphics[width=0.23cm]{Figure/mail.png}}}$ \and
Chanyoung Park$^{1, \href{mailto:cy.park@kaist.ac.kr}{\includegraphics[width=0.23cm]{Figure/mail.png}}}$
}

\authorrunning{Kim et al.}

\institute{
$^{1}$Korea Advanced Institute of Science and Technology (KAIST) \\
$^{2}$Oracle \\ 
$^{3}$University of California, San Diego \\
$^{4}$Electronics and Telecommunications Research Institute (ETRI)
}

\maketitle


\begin{abstract}
Video Large Language Models (Video LLMs) incur high inference latency due to a large number of visual tokens provided to LLMs. To address this, training-free visual token pruning has emerged as a solution to reduce computational costs; however, existing methods are primarily validated on Multiple-Choice Question Answering (MCQA) benchmarks, where coarse-grained cues often suffice. In this work, we reveal that these methods suffer a sharp performance collapse on fine-grained understanding tasks requiring precise visual grounding, such as hallucination evaluation. To explore this gap, we conduct a systematic analysis and identify \textit{sink tokens}--semantically uninformative tokens that attract excessive attention--as a key obstacle to fine-grained video understanding. When these sink tokens survive pruning, they distort the model's visual evidence and hinder fine-grained understanding.
Motivated by these insights, we propose \textbf{S}ink-\textbf{To}ken-aware \textbf{P}runing (\proposed{}), a simple yet effective plug-and-play method that introduces a sink score to quantify each token's tendency to behave as a sink and applies this score to existing spatial and temporal pruning methods to suppress them, thereby enhancing video understanding. To validate the effectiveness of~\proposed{}, we apply it to state-of-the-art pruning methods (VisionZip, FastVid, and Holitom) and evaluate it across diverse benchmarks covering hallucination, open-ended generation, compositional reasoning, and MCQA. Our results demonstrate that~\proposed{} significantly boosts performance, even when pruning up to 90\% of visual tokens. Our code is available at \href{https://github.com/rlqja1107/SToP}{https://github.com/rlqja1107/SToP}
\end{abstract}

\vspace{-4.5ex}
\section{Introduction}
\label{sec:intro}
\vspace{-1ex}
Video Large Language Models (Video LLMs)~\cite{zhang2025videollama,li2024llava,lin2024video,bai2023qwen,kim2025compodistill} have demonstrated their potential in video understanding by achieving remarkable performance in video-centric tasks, such as video question answering. However, they typically require processing thousands of visual tokens across multiple frames, leading to substantial inference latency and computational cost that hinders practical deployment.
While previous studies~\cite{li2024llama,maaz2024video,xu2024pllava,shen2024longvu} have explored visual token compression during training, these approaches often demand computationally expensive retraining. 

\begin{figure}[t]
    \centering
    \includegraphics[width=0.94\linewidth]{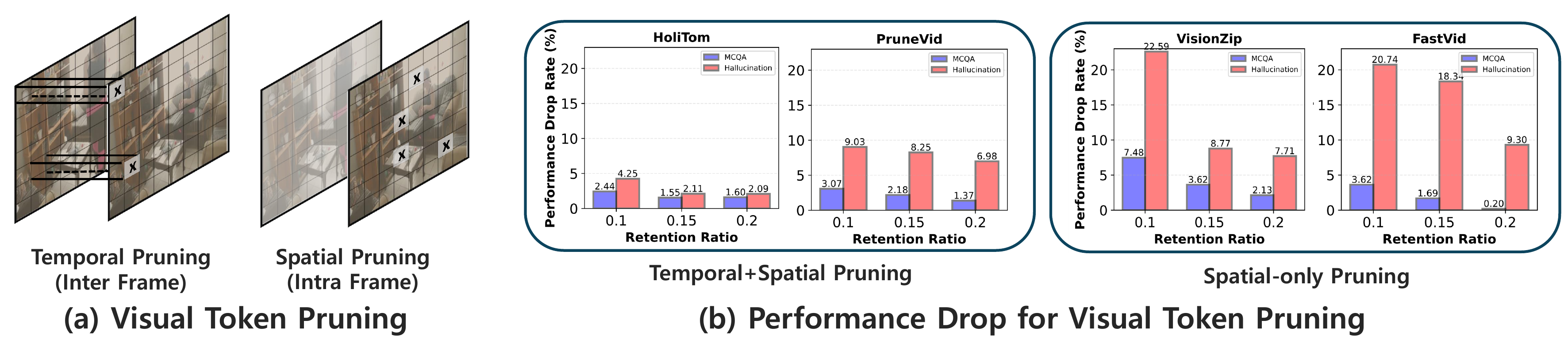}
    \vspace{-2.5ex}
    \caption{
    (a) Overview of temporal and spatial pruning. (b) Performance drop rate relative to the vanilla model for temporal+spatial pruning and spatial-only pruning methods on the MCQA (MVBench~\cite{li2024mvbench}) and hallucination benchmark (EventHallusion\cite{zhang2024eventhallusion}).}   
    \label{fig:intro}
    \vspace{-4.0ex}
\end{figure}

To circumvent this, many recent studies~\cite{yang2025visionzip,huang2025prunevid,shen2025fastvid,zhang2025beyond} adopt training-free visual token pruning, which compresses the visual tokens and reduces the number of tokens fed into the LLM without fine-tuning. In this regard, two main approaches, shown in~\cref{fig:intro}(a), are commonly employed for visual token pruning: \textbf{Temporal pruning} (inter-frame) reduces redundancy along the temporal dimension by merging similar tokens at the same patch locations across frames, while \textbf{Spatial pruning} (intra-frame) identifies and retains salient tokens within each frame based on importance metrics (e.g., attention weights), merging the pruned tokens to preserve contextual information.
Prior work on visual token pruning either combines temporal and spatial pruning (temporal+spatial pruning)~\cite{shao2025holitom,huang2025prunevid,fan2026flashvid,hyun2025multi} or mainly focuses on spatial pruning (spatial-only pruning)~\cite{zhang2025beyond,yang2025visionzip,shen2025fastvid,zhang2024cls} to improve efficiency. For example, HoliTom~\cite{shao2025holitom} performs segment-wise temporal pruning followed by attention-based spatial pruning. On the other hand, FastVid~\cite{shen2025fastvid} performs spatial pruning within each frame based on density and attention scores, without temporal pruning. Overall, these approaches facilitate significant inference acceleration by substantially reducing the visual token budget provided to the LLM.

However, despite their efficiency gains, most token pruning methods~\cite{shao2025holitom,shen2025fastvid,yang2025visionzip,zhang2024cls,zhang2025beyond} are mainly evaluated on Multiple-Choice Question Answering (MCQA) benchmarks~\cite{li2024mvbench,mangalam2023egoschema,fu2025video}, where the models can often shortcut to the correct answer using language priors and coarse visual cues~\cite{chen2024cg}, potentially masking degradations in the underlying model's video understanding.
In real-world applications, by contrast, users typically pose open-ended queries without predefined options in conversational scenarios~\cite{maaz2024video},
or require intricate spatio-temporal reasoning over open-world videos~\cite{kim2025videocomp,qiu2025step}, where correct answers must be grounded in fine-grained visual cues (e.g., subtle changes in actions and objects).
In this regard, it is crucial to examine whether existing visual token pruning methods indeed preserve such fine-grained visual cues while reducing computation.

To explore this gap, we evaluate pruning methods through the lens of hallucination\footnote{For example, prompts such as~\texttt{Please describe the video in detail} demand strong visual grounding.}, which is highly sensitive to accurate visual evidence and thus serves as a proxy for how well fine-grained visual cues are preserved. Specifically, we measure the performance drop relative to an unpruned baseline (LLaVA-OneVision-7B~\cite{li2024llava}), which uses all visual tokens over 32 frames, under varying retention ratios on a hallucination dataset (EventHallusion~\cite{zhang2024eventhallusion}) and an MCQA dataset (MVBench~\cite{li2024mvbench}).
For deeper analysis, we compare temporal+spatial pruning approach with spatial-only approach. As shown in~\cref{fig:intro}(b), we observe that while both approaches remain relatively robust on the MCQA dataset, their performance collapses much more sharply on the hallucination dataset, indicating that existing pruning methods tend to be more vulnerable on tasks requiring fine-grained visual grounding than on MCQA.
Moreover, temporal+spatial pruning is consistently more robust than spatial-only pruning, especially on the hallucination dataset, suggesting that the mechanisms underlying spatial pruning are particularly prone to hindering fine-grained video understanding.

Building on these findings, in~\cref{sec:explore_hallu}, we analyze why spatial-only pruning and temporal+spatial pruning yields different levels of video understanding, and identify a key obstacle that hinders fine-grained video understanding. Specifically, we find that a small subset of visual tokens exhibit a \textit{sink} phenomenon~\cite{xiao2023efficient,jiang2025vision}, where attention is overly concentrated on regions that contain little semantic information. If such sink tokens survive pruning, the model is more likely to hallucinate because its generation is grounded in distorted visual evidence.
Interestingly, although temporal pruning is primarily designed to remove inter-frame redundancy, it often suppresses sink tokens as a byproduct, which helps explain the greater robustness of temporal+spatial pruning.

Motivated by our analysis, we propose \textbf{S}ink-\textbf{To}ken-aware \textbf{P}runing (\proposed{}), a training-free visual token pruning method that explicitly targets sink tokens during pruning. Specifically, we first define a \textit{sink score} that quantifies each token's tendency to act as a sink. Based on this score, we design a Sink-Token-aware Spatial Pruning (STSP) module that adjusts the attention distribution of sink tokens, lowering their priority for retention—even when conventional attention-based importance scores would otherwise rank them highly. This design explicitly encourages the pruning of tokens that may appear salient under standard attention-based criteria but in fact correspond to sink behavior. In addition, we design a Sink-Token-aware Temporal Pruning (STTP) module, which further promotes the elimination of sink tokens along the time dimension.

To validate the effectiveness of~\proposed{}, we apply it to the state-of-the-art pruning methods (VisionZip~\cite{yang2025visionzip}, FastVid~\cite{shen2025fastvid}, and HoliTom~\cite{shao2025holitom}) and evaluate across multiple aspects of video understanding: hallucination evaluation (EventHallusion~\cite{zhang2024eventhallusion}), compositional reasoning (VideoComp~\cite{kim2025videocomp}), and open-ended generation (VCG-Bench~\cite{maaz2024video}). Our results demonstrate that~\proposed{} substantially enhances video understanding even under an extreme token budget (e.g., pruning 90\% of tokens). These benefits also carry over to MCQA benchmarks, where we observe consistent performance gains.

We summarize our main contributions as follows: 1) Through a systematic analysis of visual token pruning, we identify that \textit{sink tokens} are a key obstacle that hinders fine-grained video understanding. 2) We propose~\proposed{}, a training-free token pruning method that introduces a sink score and incorporates it into both spatial and temporal pruning to explicitly guide the pruning of sink tokens. 3) Extensive experiments across diverse video understanding tasks (e.g., hallucination, compositional reasoning, open-ended generation, and MCQA) demonstrate the effectiveness of~\proposed{}, consistently outperforming baselines.


\section{Preliminary}
\label{sec:preliminaries}

\vspace{-1ex}
\subsection{Inference Complexity of Video LLMs}
\noindent \textbf{Architecture of Video LLMs. } Given a user's question $q$ and an input video of $T$ frames, a visual encoder (e.g., SigLIP~\cite{zhai2023sigmoid}) processes $T$ frames independently to extract $n_v$ patch embeddings per frame, followed by projecting them into the language embedding space with a neural network (e.g., MLP~\cite{li2024llava,lin2024vila}). This yields visual tokens $H_v\in \mathbb{R}^{T\times n_v\times d}$ , where $n_v$ is the number of visual tokens per frame and $d$ is the dimensionality of the LLM word embeddings.  The user's question prompt is converted into text tokens $H_q\in \mathbb{R}^{n_q\times d}$, where $n_q$ is the number of text tokens for $q$. 
The model then concatenates $H_v$ and $H_q$ and feeds the combined sequence into the LLM, which generates the response autoregressively by predicting the next text token at each decoding step.

\smallskip
\noindent \textbf{Computational Complexity.} We analyze the inference cost of Video LLMs in terms of Floating-Point Operations (FLOPs) and how the number of frames and visual tokens affects the overall computation. Given that the dominant cost arises from self-attention and feed-forward network (FFN) within the LLM, the total FLOPs across the LLM's $L$ layers can be expressed as:

\begin{equation}
\text{FLOPs}=L\times (4nd^2+2n^2d+2ndm),
\label{eqn:1}
\end{equation}
where $n=T\cdot n_v+n_q$ is the input sequence length and $m$ is the intermediate dimension of the FFN. It is important to note that the term $T\cdot n_v$ typically dominates $n$ (e.g., $n_v=6,270$ for 32 frames on LLaVA-OneVision~\cite{li2024llava}, while $n_q\approx 20$). Furthermore, adding frames increases the sequence length by $n_v$ per frame, incurring the attention cost quadratically through the $2n^2d$ term and the projection and FFN costs linearly through the $4nd^2 + 2ndm$ terms, with all costs further scaled by the depth $L$. 
This analysis highlights the importance of reducing the number of visual tokens $T \cdot n_v$ for efficient inference. In~\cref{sec:main_result}, we show that~\proposed{} remains robust on fine-grained video tasks (e.g., hallucination) even when pruning 90\% of visual tokens (i.e., retaining only $0.1 \cdot T \cdot n_v$).

\vspace{-1ex}
\subsection{Token Pruning Approach}
To reduce the number of visual tokens provided to LLMs, spatial~\cite{yang2025visionzip,shen2025fastvid,wang2024cls} and temporal~\cite{tao2025dycoke,fan2026flashvid,shao2025holitom} pruning approaches are commonly used.

\smallskip
\noindent \textbf{Spatial Pruning.} Within each frame, it is important to retain the salient visual tokens while removing those that carry little semantic information. To this end, many token pruning methods~\cite{fan2026flashvid,yang2025visionzip,shao2025holitom,shang2025llava,zhang2024cls,wang2024cls} utilize the attention weights of a vision encoder, retaining tokens with high attention scores. Specifically, following prior work~\cite{zhang2024cls,shen2025fastvid,shao2025holitom,yang2025visionzip,wang2024cls}, we perform token selection using the CLS token. For visual encoders without an explicit CLS token (e.g., SigLIP~\cite{zhai2023sigmoid}), we instead construct a CLS-equivalent attention~\cite{shao2025holitom,fan2026flashvid,yang2025visionzip}. More precisely, the attention matrix is computed as:
\begin{equation}
\hat{A}=\text{Softmax}(QK^T / \sqrt{d}) \in \mathbb{R}^{T\times n_v \times n_v},
\label{eqn:2}
\end{equation}
where $Q$ and $K$ are the query and key of the visual tokens, respectively. Following~\cite{fan2026flashvid,shao2025holitom,yang2025visionzip}, a token's attention score is obtained by averaging the attention matrix column-wise, indicating that a token receiving more attention from other tokens is deemed more important. This yields an attention score matrix $A \in \mathbb{R}^{T\times n_v}$, which is used to select salient visual tokens with high attention score for each frame.

\smallskip
\noindent \textbf{Temporal Pruning.} 
In a video, some regions may stay unchanged over consecutive frames (e.g., a stationary object), resulting in temporal redundancy. To address it, prior work~\cite{huang2025prunevid,shao2025holitom,hyun2025multi,tao2025dycoke} commonly detects redundancy by computing the similarity between tokens at the same patch location in adjacent frames, and then prunes tokens whose similarity exceeds a threshold. Formally, the patch indices pruned at frame $t$ are defined as:
\begin{equation}
P_t = \{i|\text{sim}(H_v^{i,t}, H_v^{i,t+1}) > \tau, i \in \left[ 1,n_v \right] \},
\label{eqn:3}
\end{equation}
where $\tau$ is pruning threshold, $H_v^{i,t}$ denotes the token at the $i$-th patch location in the $t$-th frame, and $\text{sim}(\cdot, \cdot)$ is the cosine similarity between the two tokens. In practice, this pairwise similarity is often computed for every adjacent-frame pair within a temporally segmented clip and then aggregated (e.g., product); if the aggregated similarity exceeds a threshold, the tokens at the corresponding location within the clip are removed~\cite{tao2025dycoke,shao2025holitom,huang2025prunevid}.

\section{Identifying a Key Obstacle to Fine-Grained Video Understanding}
\label{sec:explore_hallu}

In this section, we examine the gap in video understanding between temporal+spatial pruning and spatial-only pruning approaches, as shown in~\cref{fig:intro}(b), and aim to identify a key obstacle to fine-grained video understanding. In line with experimental settings in~\cref{sec:intro}, we conduct this study on the EventHallusion~\cite{zhang2024eventhallusion} dataset using LLaVA-OneVision~\cite{li2024llava}.

\begin{figure}[t]
    \centering
    \includegraphics[width=0.9\linewidth]{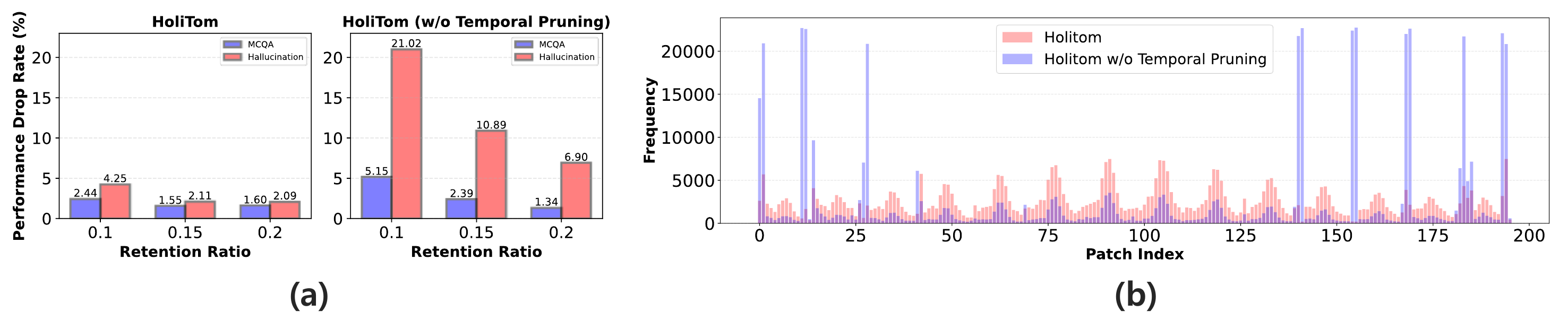}
    \vspace{-3.5ex}
    \caption{
    (a) Performance degradation upon the removal of temporal pruning. (b) Comparison of selected visual token distributions between temporal+spatial pruning and the variant without temporal pruning as a retention ratio of 0.1. We use the EventHallusion~\cite{zhang2024eventhallusion} dataset for this analysis.}   
    \label{fig:holitom}
    \vspace{-3.5ex}
\end{figure}

\subsection{Impact of Temporal Pruning for Fine-Grained Understanding}
\label{sec:temporal_pruning}

We hypothesize that the temporal pruning component helps preserve fine-grained visual cues beyond merely reducing temporal redundancy, as it is the core difference between the two approaches.
To test this, we utilize Holitom~\cite{shao2025holitom}, a temporal+spatial pruning method, as our baseline and evaluate the performance drop when we intentionally remove temporal pruning.\footnote{To maintain a constant token budget when temporal pruning is removed, we proportionally increase the number of tokens selected during spatial pruning.}. As shown in~\cref{fig:holitom}(a), we observe that removing temporal pruning substantially increases hallucinations, indicating that temporal pruning supports fine-grained video understanding, which we further discuss in~\cref{sec:sink_token_obstacle}. To uncover the cause of this drop, we analyze the distribution of selected visual tokens over the entire video. As shown in~\cref{fig:holitom}(b), we observe that without temporal pruning, a small subset of visual tokens is selected with disproportionately high frequency. This suggests that repeatedly selecting these tokens may harm fine-grained video understanding.
Before validating it, we trace the source of these abnormal selection frequencies.

\label{sec:origin_high_frequency}
\begin{wrapfigure}{r}{0.55\linewidth}
  \centering
  \vspace{-25pt} 
  \includegraphics[width=0.99\linewidth]{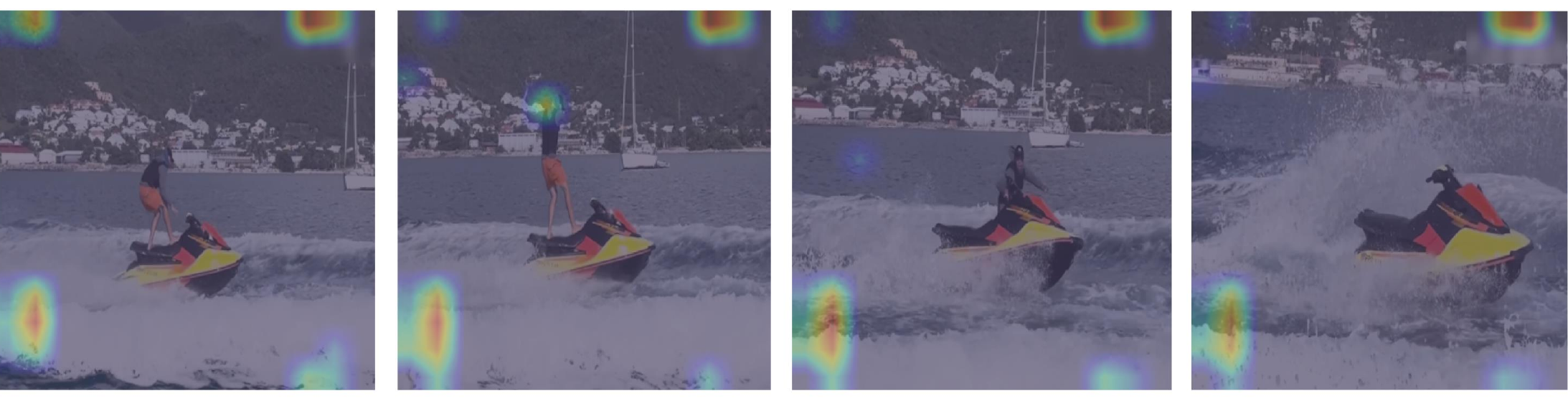}
  \vspace{-5ex}
  \caption{Visualization of attention scores across consecutive frames.}
  \label{fig:example_attn_map}
  \vspace{-21pt}
\end{wrapfigure}

\vspace{-1ex}
\subsection{Source of High Frequency: Spatially Persistent High Attention}
\label{sec:source_high_freq}
To characterize these high-frequency tokens identified in~\cref{fig:holitom}(b), we visualize the patch-wise attention scores over consecutive frames. As shown in~\cref{fig:example_attn_map},
we observe that high attention scores persist at fixed spatial coordinates throughout the temporal sequence. Moreover, given that salient objects tend to appear near the center of the frame~\cite{tatler2007central}, these persistent high-attention regions are likely to correspond to semantically sparse background areas. Drawing a parallel in LLMs, where attention highly concentrates on semantically uninformative tokens at fixed positions (e.g., the BOS token)~\cite{xiao2023efficient,sun2024massive}, we refer to these spatially persistent high-attention tokens as \textit{\textbf{sink tokens}}\footnote{Regarding a discussion of differences in perspectives on sink tokens in LLMs, please refer to Appendix~\ref{app_sec:sink_diff}.}. Due to their high attention weights, sink tokens are preferentially retained during spatial pruning. In other words, their repeated retention at fixed coordinates leads to abnormally high selection frequencies (e.g., the lower-left patches in~\cref{fig:example_attn_map} with indices 154 and 155), observed in~\cref{fig:holitom}(b).

\vspace{-2ex}
\subsection{\textit{Sink Tokens} as an Obstacle to Fine-Grained Understanding}
\label{sec:sink_token_obstacle}
\vspace{-1ex}
\begin{wrapfigure}{r}{0.36\linewidth}
  \centering
  \vspace{-26pt} 
  \includegraphics[width=0.99\linewidth]{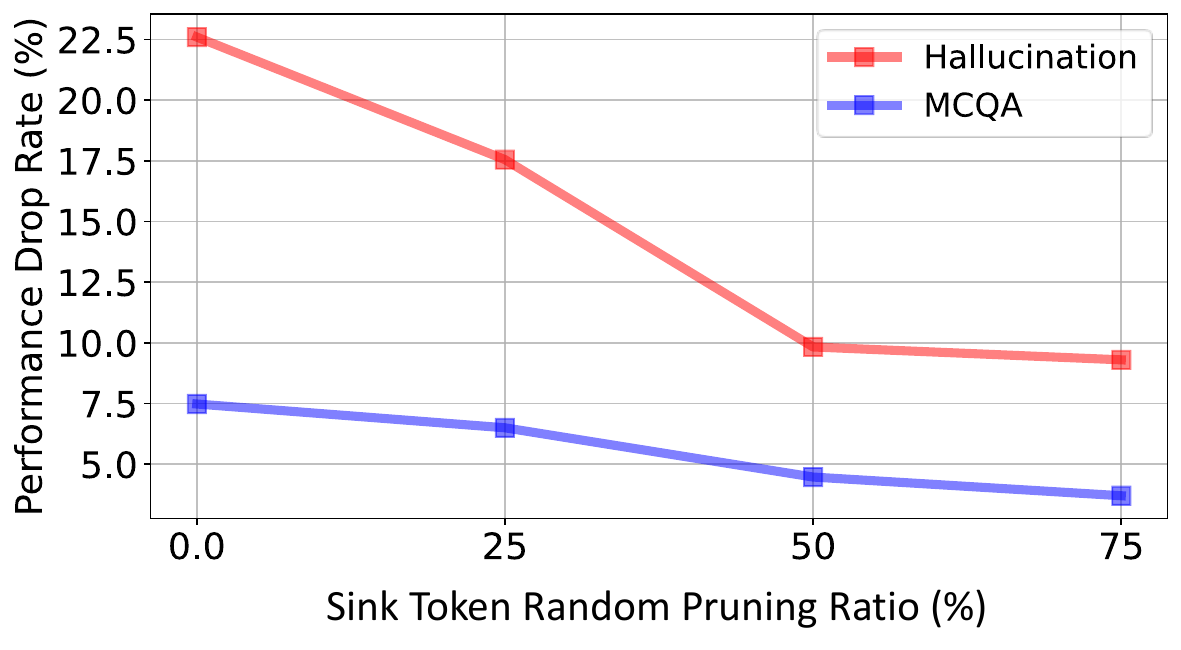}
  \vspace{-5ex}
  \caption{Performance degradation rate across varying sink token removal ratios.}
  \label{fig:naive_approach}
  \vspace{-21pt}
\end{wrapfigure}
To verify whether the frequent selection of sink tokens directly hinders fine-grained video understanding and exacerbates hallucination, we perform a diagnostic experiment using VisionZip~\cite{yang2025visionzip}, a spatial-only pruning method that is relatively prone to hallucination (see~\cref{fig:intro}). Specifically, to deliberately reduce the retention of sink tokens in a simple manner, we randomly remove sink tokens at varying ratios from the selected token set\footnote{To maintain a constant token budget, removed sink tokens are replaced with an equal number of tokens with the next highest attention scores.}. As shown in~\cref{fig:naive_approach}, we observe that as the removal ratio of sink tokens increases, performance degradation in hallucination diminishes significantly while improving MCQA performance. This indicates that sink tokens are directly detrimental to fine-grained video understanding. We hypothesize this is because sink tokens generally carry limited semantic information despite high attention scores. As a result, repeatedly retaining them consumes the limited token budget, leaving insufficient capacity for truly salient visual tokens and causing text generation based on distorted or incomplete evidence. Regarding a detailed ablation on the inherent impact of sink tokens, please refer to~\cref{app_sec:ablation_sink_token}.

\smallskip
\noindent \textbf{Rethinking the Role of Temporal Pruning.} While temporal pruning is originally designed to reduce temporal redundancy~\cite{shao2025holitom,huang2025prunevid,fan2026flashvid}, we posit that it implicitly acts as a sink token suppressor. This is because sink tokens tend to reside in background regions, as discussed in~\cref{sec:origin_high_frequency}, 
making their representations highly similar across adjacent frames and thus more likely to be removed by temporal pruning.
In fact, when comparing the number of sink tokens\footnote{Herein, the set of sink tokens is defined as the top 10\% highest-frequency tokens in~\cref{fig:holitom}(b).} selected by Holitom (temporal+spatial pruning) against an variant without temporal pruning (spatial pruning), which is introduced in~\cref{sec:temporal_pruning}, we observe that temporal pruning reduces sink tokens by 83\% ($384 \rightarrow 66$). Coupled with the performance drop observed upon the removal of temporal pruning (\cref{fig:holitom}(a)), this correlation further supports our hypothesis that the frequent selection of sink tokens is a key obstacle to fine-grained video understanding.

\section{Proposed Method: Sink-Token-aware Pruning (\proposed{})}
\label{sec:method}
\smallskip
\begin{wrapfigure}{r}{0.59\linewidth}
  \centering
  \vspace{-26pt} 
  \includegraphics[width=0.99\linewidth]{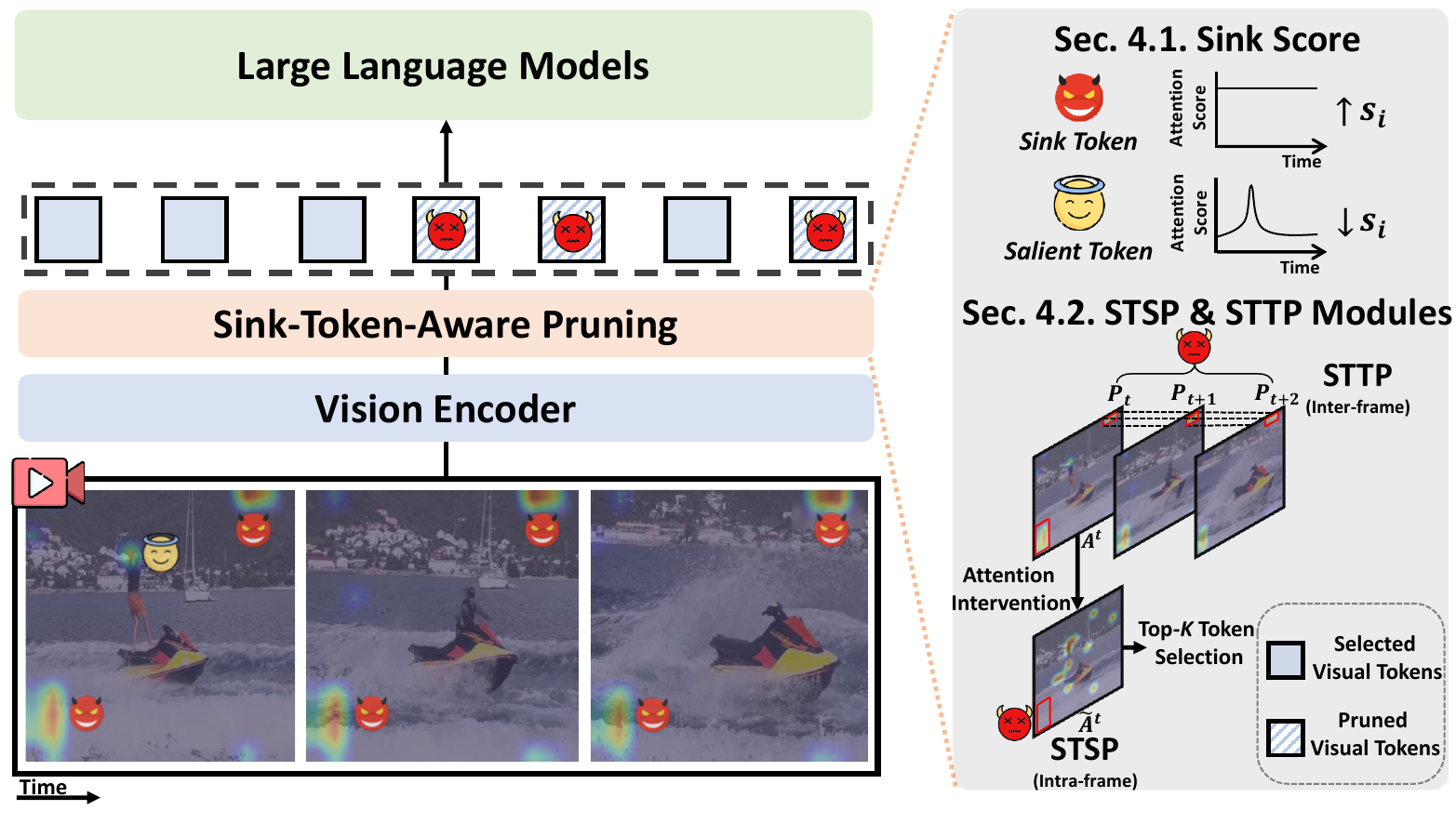}
  \vspace{-6ex}
  \caption{Overall Framework.}
  \label{fig:main_figure}
  \vspace{-21pt}
\end{wrapfigure}

Building on our analysis in~\cref{sec:explore_hallu}, we propose~\proposed{}, a plug-and-play token pruning method designed to explicitly guide the pruning of sink tokens. We begin by defining a sink score to quantify how likely each token is to behave as a sink token (\cref{sec:temporal_sink}). This score is then seamlessly incorporated into existing spatial and temporal frameworks~\cite{yang2025visionzip,shao2025holitom,shen2025fastvid,huang2025prunevid} via two proposed modules: the STSP module for spatial pruning and the STTP module for temporal pruning (\cref{sec:sink_spatial_prune}).

\vspace{-1ex}
\subsection{Quantifying Token Persistence via Sink Score}
\label{sec:temporal_sink}
While sink tokens can be broadly identified by the distribution gap between the temporal+spatial and spatial-only pruning (\cref{fig:holitom}(b)), it remains challenging to finely distinguish sink tokens from truly salient tokens, as both exhibit high attention values. However, as discussed in~\cref{sec:source_high_freq}, sink tokens are typically characterized by spatially persistent high attention across the temporal dimension. To formalize this behavior, we define the sink score $s_i$ for each visual token $i$ as:
\vspace{-2ex}
\begin{equation}
s_i=\text{MinMax-Norm}(\hat{s}_i^{w}), \hat{s}_i=\sum_{t=1}^{T}A_i^t,
\label{eqn:4}
\end{equation}
where $A_i^t$ denotes the attention score of the $i$-th token in the $t$-th frame ($i\in\left[ 1, n_v \right]$). Intuitively, $\hat{s}_i$ yields a high value when a token's attention remains consistently elevated across frames, identifying it as a likely sink token. On the other hand, non-sink tokens that are only transiently salient—scoring highly in only a few frames—result in a lower $\hat{s}_i$. Lastly, we apply min-max normalization (\text{MinMax-Norm}) to the power term $\hat{s}_i^{w}$ to map the values to the range $\left[ 0,1 \right]$. The hyperparameter $w$ serves to sharpen the distribution of $s_i$, suppressing the influence to the non-sink tokens while effectively highlighting the sink tokens. Regarding a detailed analysis of $w$, please refer to Appendix~\ref{app_sec:dist_w}.

\subsection{Sink-Token-aware Pruning}
\label{sec:sink_spatial_prune}
\noindent \textbf{STSP: Sink-Token-aware Spatial Pruning.} Leveraging the sink score $s_i$, we propose a plug-and-play spatial pruning method, termed \textbf{STSP} module, which explicitly attenuates the selection of sink tokens during pruning. Conventional spatial pruning methods~\cite{yang2025visionzip,shen2025fastvid,shao2025holitom} generally retain tokens with high attention scores, operating under the assumption that these scores directly correlate with visual importance. However, their methods often inadvertently select sink tokens—which exhibit high attention values despite a lack of semantics—thereby hindering fine-grained video understanding. To address this, we integrate the sink score into the attention mechanism as:
\begin{equation}
\tilde{A}_i^t=A_i^t-\mu_s \cdot s_i,
\label{eqn:5}
\end{equation}
where $\mu_s$ is a hyperparameter controlling the influence of the sink score. By penalizing sink-prone tokens with high sink scores, we effectively reduce sink tokens' attention score, preventing the selection of semantically sparse tokens that would otherwise dominate the selection process. This intervention ensures that a higher proportion of genuinely salient tokens are selected, ultimately enhancing video understanding even at a low retention ratio. 

\smallskip
\noindent \textbf{STTP: Sink-Token-aware Temporal Pruning.} Although temporal pruning implicitly discourages the retention of sink tokens, these tokens still survive after temporal pruning, as discussed in~\cref{sec:sink_token_obstacle}. To mitigate this, we propose \textbf{STTP} module, which explicitly leverages the sink score to further facilitate the pruning of sink tokens along the temporal axis. Formally, we integrate this score into~\cref{eqn:3} as follows:
\begin{equation}
P_t = \{i|\text{sim}(H_v^{i,t}, H_v^{i,t+1}) + \mu_t \cdot s_i > \tau , i \in \left[ 1,n_v \right] \},
\label{eqn:6}
\end{equation}
where $\mu_t$ is the hyperparameter to control the weight of $s_i$. The term $\mu_t \cdot s_i$ acts as a factor that pushes sink tokens above the threshold $\tau$, thereby explicitly assigning them to the pruning set $P_t$.


\section{Experiment}
\subsection{Experimental Setting}
\label{sec:exp_setting}
\noindent \textbf{Dataset.} 
To assess fine-grained visual grounding, we evaluate~\proposed{} across multiple dimensions of video understanding. Specifically, we utilize EventHallusion~\cite{zhang2024eventhallusion} to measure \textit{hallucination} via binary questions (Binary), which probe susceptibility to language bias and therefore require fine-grained video understanding—and open-ended descriptive tasks (Desc.). We utilize VideoComp~\cite{kim2025videocomp} to evaluate \textit{compositional reasoning}, specifically focusing on complex temporal and structural relationships and subtle action transitions, and report results on two domains: ActivityNet (Act.) and YouCook2 (YC).
Finally, to evaluate \textit{open-ended generation}, we employ VCG-Bench~\cite{maaz2024video}. 
For all free-form text generation evaluations (EvetHallusion-Desc., and VCG-Bench), we use GPT-5~\cite{singh2025openai} for scoring. Furthermore, we evaluate~\proposed{} on five widely-used MCQA benchmarks: MVBench~\cite{li2024mvbench}, VideoMME~\cite{fu2025video}, NextQA~\cite{xiao2021next}, LongVideoBench~\cite{wu2407longvideobench}, and MLVU~\cite{zhou2025mlvu}. Additional details on these datasets and their metrics are described in Appendix~\ref{app_sec:dataset_detail}.

\smallskip
\noindent \textbf{Baselines.}
\proposed{} is agnostic to attention-based spatial pruning, allowing it to integrate seamlessly with various pruning methods. Specifically, we apply it to spatial pruning (VisionZip~\cite{yang2025visionzip} FastVid~\cite{shen2025fastvid}) equipped with the STSP module, and temporal+spatial pruning (Holitom~\cite{shao2025holitom}) equipped with the STSP and STTP modules. We compare these integrated versions against their respective baselines, and the state-of-the-art method, PruneVid~\cite{huang2025prunevid}. To further demonstrate the adaptability of~\proposed{}, we apply it to FlashVid~\cite{fan2026flashvid}, the most recent visual token pruning method, described in Appendix~\ref{app_sec:flash_vid}.

\smallskip
\noindent \textbf{Implementation Details.}
Following the experimental setup~\cite{shao2025holitom}, we implement our method on LLaVA-OneVision-7B~\cite{li2024llava} and LLaVA-Video-7B~\cite{zhang2024llava}. For all evaluations, we utilize 32 frames, resulting in a total of $32\times 196$ visual tokens. In line with~\cite{shao2025holitom,fan2026flashvid}, we evaluate our method using retention ratios of 10\%, 15\%, and 20\%.
Regarding hyperparameters, we set $w$ to 1.1 across all experiments. The threshold parameter $\mu_s$ is set to 0.03 for VisionZip and 0.02 for FastVid.  All models are evaluated on an NVIDIA GeForce A6000 48GB GPU. Unless otherwise stated, we conduct experiments using LLaVA-OneVision-7B as the primary backbone. Additional implementation details are provided in the Appendix~\ref{app_sec:implementation}. Regarding experiments of hyperparameter sensitivity, please refer to Appendix~\ref{app_sec:hyper}.

\begin{table}[t]
\centering
\caption{Performance comparison on benchmarks requiring fine-grained video understanding. Performance Drop Rate denotes the macro-average of performance degradation relative to the Vanilla model across all categories. Desc., Consist., Temp., Act., YC are Description, Consistency, Temporal, ActivityNet, and YouCook2, respectively.}
\vspace{-1ex}
\setlength{\tabcolsep}{2.5pt}
\resizebox{.95\linewidth}{!}{
\begin{tabular}{l|c|cc|cc|ccccc|c}
\toprule
\multicolumn{1}{c|}{\multirow{2}{*}{\textbf{Methods}}} & \multirow{2}{*}{\textbf{\begin{tabular}[c]{@{}c@{}}Retention\\ Ratio\end{tabular}}} & \multicolumn{2}{c|}{\textbf{EventHallusion}} & \multicolumn{2}{c|}{\textbf{VideoComp}} & \multicolumn{5}{c|}{\textbf{VCG-Bench}}             & \multirow{2}{*}{\textbf{\begin{tabular}[c]{@{}c@{}}Performance \\ Drop Rate\end{tabular}}} \\
\multicolumn{1}{c|}{}                                  &                                                                                     & Binary                & Desc.                & Act.                & YC               & Consist. & Temp. & Context & Detail & Correct &                                     \\ \midrule \midrule
Vanilla                                                & 100\%                                                                               & 63.33                 & 38.74                & 70.06              & 70.95              & 3.32        & 2.30     & 3.28    & 2.38   & 2.98    & -                                   \\ \midrule
PruneVid                                               & \multirow{7}{*}{20\%}                                                               & 60.04                 & 32.45                & 70.58              & 70.52              & 3.16        & 2.15     & 3.12    & 2.14   & 2.86    & 5.74\%                              \\
Holitom                                               &                                                                                     & 63.57                 & 35.76                & 70.78              & 70.95              &     3.22        &  2.15        &   3.20      &   2.22     &   2.95     &   2.41\%                                  \\
\rowcolor{gainsboro}Holitom+\proposed{}                &     \cellcolor{white}                                                                                  & 63.57                 & 38.41                & 71.04              & 70.66              &          3.23   &  2.18        &   3.21      &   2.23     &  2.96       &   $\text{1.84\%}_{\textcolor{blue}{\downarrow 0.58\%}}$                                  \\
FastVid                                                &                                                                                     & 55.99                 & 37.09                & 60.72              & 59.81              & 3.15        & 2.11     & 3.16    & 2.16   & 2.90    & 8.21\%                              \\
\rowcolor{gainsboro}FastVid+\proposed{}                &   \cellcolor{white}                                                                                    & 62.59                 & 37.75                & 70.30              & 71.12              & 3.15        & 2.15     & 3.21    & 2.24   & 2.96    & $\text{2.61\%}_{\textcolor{blue}{\downarrow 5.60\%}}$                              \\
VisionZip                                            &                                                                                     & 58.44                 & 35.76                & 69.46              & 69.49              & 3.22        & 2.17     & 3.14    & 2.14   & 2.91    & 4.85\%                              \\
\rowcolor{gainsboro}VisionZip+\proposed{}              &   \cellcolor{white}                                                                                    & 62.59                 & 37.42                & 70.18              & 70.78              & 3.16        & 2.09     & 3.17    & 2.19   & 2.89    & $\text{3.66\%}_{\textcolor{blue}{\downarrow 1.19\%}}$                              \\ \midrule
PruneVid                                               & \multirow{7}{*}{15\%}                                                               & 58.44                 & 35.10                & 69.58              & 70.18              & 3.14        & 2.17     & 3.08    & 2.11   & 2.87    & 5.68\%                              \\
Holitom                                                &                                                                                     & 63.08                 & 36.42                & 69.82              & 70.01              & 3.18        & 2.15     & 3.17    & 2.16   & 2.92    & 3.52\%                              \\
\rowcolor{gainsboro}Holitom+\proposed{}                &    \cellcolor{white}                                                                                  & 64.06                 & 37.42                & 70.18              & 70.61              & 3.15        & 2.16     & 3.18    & 2.20   & 2.96    & $\text{2.78\%}_{\textcolor{blue}{\downarrow 0.74\%}}$                              \\
FastVid                                                &                                                                                     & 52.57                 & 30.46                & 60.00              & 59.04              & 3.14        & 2.10     & 3.03    & 2.06   & 2.80    & 12.30\%                             \\
\rowcolor{gainsboro}FastVid+\proposed{}                &   \cellcolor{white}                                                                                    & 60.88                 & 40.73                & 69.46              & 70.61              & 3.15        & 2.09     & 3.14    & 2.17   & 2.88    & $\text{3.42\%}_{\textcolor{blue}{\downarrow 8.89\%}}$                              \\
VisionZip                                              &                                                                                     & 58.44                 & 34.44                & 67.78              & 68.29              & 3.05        & 2.10     & 3.03    & 2.05   & 2.78    & 7.87\%                              \\
\rowcolor{gainsboro}VisionZip+\proposed{}              &     \cellcolor{white}                                                                                  & 61.61                 & 38.08                & 68.38              & 69.84              & 3.16        & 2.12     & 3.12    & 2.15   & 2.87    & $\text{4.36\%}_{\textcolor{blue}{\downarrow 3.51\%}}$                              \\ \midrule
PruneVid                                               & \multirow{7}{*}{10\%}                                                               & 56.72                 & 32.12                & 68.26              & 68.38              & 3.02        & 2.04     & 3.02    & 2.04   & 2.78    & 9.22\%                              \\
Holitom                                                &                                                                                     & 60.88                 & 36.75                & 68.38              & 68.98              & 3.07        & 2.06     & 3.08    & 2.07   & 2.84    & 6.22\%                              \\
\rowcolor{gainsboro}Holitom+\proposed{}                &    \cellcolor{white}                                                                                   & 62.59                 & 39.74                &        69.24      &      69.10         & 3.07        & 2.03     & 3.10    & 2.11   & 2.84    &      $\text{4.80\%}_{\textcolor{blue}{\downarrow 1.42\%}}$                          \\
FastVid                                                &                                                                                     & 49.63                 & 31.46                & 57.37              & 58.10              & 2.95        & 1.95     & 2.92    & 1.94   & 2.72    & 15.69\%                             \\
\rowcolor{gainsboro}FastVid+\proposed{}                &    \cellcolor{white}                                                                                   & 60.15                 & 36.75                & 68.38              & 69.75              & 3.09        & 2.03     & 3.06    & 2.09   & 2.83    &  $\text{6.32\%}_{\textcolor{blue}{\downarrow 9.37\%}}$                              \\
VisionZip                                              &                                                                                     & 50.37                 & 28.15                & 65.39              & 64.61              & 2.81        & 1.84     & 2.78    & 1.84   & 2.55    & 16.79\%                             \\
\rowcolor{gainsboro}VisionZip+\proposed{}              &             \cellcolor{white}                                                                        & 60.39                 & 36.09                & 67.78              & 68.89              & 3.05        & 2.06     & 3.05    & 2.08   & 2.80    & $\text{6.87\%}_{\textcolor{blue}{\downarrow 9.92\%}}$                              \\ \bottomrule
\end{tabular}
}
\label{tab:fine_main}
\vspace{-2ex}
\end{table}

\begin{table}[t]
\centering
\caption{Performance comparison on multiple-choice question answering (MCQA) benchmarks. The Performance Drop Rate is computed as in~\cref{tab:fine_main}.}
\vspace{-2ex}
\setlength{\tabcolsep}{2.5pt}
\resizebox{.95\linewidth}{!}{
\begin{tabular}{l|c|ccccccc|c}
\toprule
\multirow{2}{*}{\textbf{Method}}   & \multirow{2}{*}{\textbf{Retention}} & \multirow{2}{*}{\textbf{MVBench}} & \multicolumn{3}{c}{\textbf{VideoMME}}                                   & \multicolumn{1}{l}{\multirow{2}{*}{\textbf{NextQA}}} & \multicolumn{1}{l}{\multirow{2}{*}{\textbf{LongVideoBench}}} & \multirow{2}{*}{\textbf{MLVU}} & \multirow{2}{*}{\textbf{\begin{tabular}[c]{@{}c@{}}Performance \\ Drop Rate\end{tabular}}} \\
                                   &                                     &                                   & Short                     & Medium               & Long                 & \multicolumn{1}{l}{}                                 & \multicolumn{1}{l}{}                                         &                                &                                                                                            \\ \midrule \midrule
Vanilla                            & 100\%                               & 58.54                             & 71.22                     & 56.22                & 48.61                & 79.64                                                & 54.23                                                        & 64.66                          & -                                                                                          \\ \midrule
PruneVid                           & \multirow{7}{*}{15\%}               & 57.26                             & 66.78                     & 54.56                & 47.22                & 79.54                                                & 54.90                                                        & 62.65                          & 2.32\%                                                                                     \\
Holitom                            &                                     & 57.63                             & 67.89                     & 56.78                & 48.00                & 79.74                                                & 54.30                                                        & 62.74                          & 1.31\%                                                                                     \\
\rowcolor{gainsboro}Holitom+\proposed{}  &  \cellcolor{white}                                     & 57.84              &   67.89    &  55.67 & 48.89 & 80.38                                 &  54.30                                        &           63.75                    &     $\text{0.95\%}_{\textcolor{blue}{\downarrow 0.37\%}}$                                                                                       \\
FastVid                            &                                     & 57.55                             & 65.89                     & 53.56                & 48.11                & 79.24                                                & 52.43                                                        & 62.55                          & 3.15\%                                                                                     \\
\rowcolor{gainsboro}FastVid+\proposed{}   &    \cellcolor{white}                                 & 57.76                             & \multicolumn{1}{l}{67.67} & 54.22                & 49.11                & 80.30                                                & 55.12                                                        & 63.43                          & $\text{1.18\%}_{\textcolor{blue}{\downarrow 1.96\%}}$                                                                                     \\
VisionZip                          &                                     & 56.42                             & 65.11                     & 54.56                & 48.11                & 79.10                                                & 52.80                                                        & 62.65                          & 3.23\%                                                                                     \\
\rowcolor{gainsboro}VisionZip+\proposed{} &    \cellcolor{white}                                 & 57.71                             & 67.33                     & 54.89                & 49.33                & 79.92                                                & 53.63                                                        & 62.93                          & $\text{1.60\%}_{\textcolor{blue}{\downarrow 1.63\%}}$                                                                                    \\ \midrule
PruneVid                           & \multirow{7}{*}{10\%}               & 56.74                             & 64.89                     & 54.22                & 47.67                & 79.10                                                 & 54.45                                                        & 61.78                          & 3.17\%                                                                                     \\
Holitom                            &                                     & 57.11                             & 64.78                     & 54.44                & 47.78                & 79.48                                                & 53.55                                                        & 61.64                          & 3.21\%                                                                                     \\
\rowcolor{gainsboro}Holitom+\proposed{}  &         \cellcolor{white}                            &   57.45                                &  66.67      &   55.22                   &         48.89             &       79.86                                               &                                                   53.33           &    62.42                            &     $\text{2.02\%}_{\textcolor{blue}{\downarrow 1.20\%}}$                                                                            \\
FastVid                            &                                     & 56.42                             & 62.55                     & 52.67                & 46.67                & 78.68                                                & 52.58                                                        & 61.09                          & 5.12\%                                                                                     \\
\rowcolor{gainsboro}FastVid+\proposed{}   &    \cellcolor{white}                                 & 57.79                             & \multicolumn{1}{l}{65.22} & 54.22                & 48.22                & 79.44                                                & 53.55                                                        & 61.59                          & $\text{2.90\%}_{\textcolor{blue}{\downarrow 2.22\%}}$                                                                                     \\
VisionZip                          &                                     & 54.16                             & 61.00                     & 52.89                & 45.78                & 77.40                                                & 49.29                                                        & 60.76                          & 7.36\%                                                                                     \\
\rowcolor{gainsboro}VisionZip+\proposed{} &    \cellcolor{white}                                 & 57.34                             & 65.67                     & 53.78                & 47.67                & 79.72                                                & 52.51                                                        & 61.96                          & $\text{3.34\%}_{\textcolor{blue}{\downarrow 4.02\%}}$                                                                                     \\ \bottomrule
\end{tabular}
}
\label{tab:mcqa_main}
\vspace{-3ex}
\end{table}

\subsection{Main Results}
\label{sec:main_result}
We evaluate~\proposed{} by integrating it into three state-of-the-art methods: VisionZip, FastVid (spatial pruning), and Holitom (temporal+spatial pruning). Our evaluation spans fine-grained video tasks and MCQA benchmarks. 

\smallskip
\noindent \textbf{Results on Fine-Grained Video tasks. }  \cref{tab:fine_main} shows the performance of~\proposed{} compared to baselines on fine-grained video tasks across the EventHallusion, VideoComp, and VCG-Bench datasets. We have the following observations: 
\textbf{1)} Existing token pruning methods suffer from significant performance drops as the retention ratio decreases. This suggests an inability to preserve fine-grained visual cues under strict token budgets. We attribute this to the fact that when the token budget is tighter, sink tokens with low semantics tend to dominate the selection process, thereby crowding out informative tokens essential for fine-grained video understanding.
\textbf{2)} Temporal+spatial pruning methods (Holitom, PruneVid) are generally more robust than spatial pruning methods (VisionZip, FastVid), indicating that, as discussed in~\cref{sec:sink_token_obstacle}, temporal pruning acts as the primary driver by implicitly suppressing sink tokens.
\textbf{3)} Applying~\proposed{} to existing frameworks (VisionZip, FastVid, Holitom) significantly mitigates performance degradation compared to their original counterparts, especially at a low retention ratio of 10\% and 15\%. This demonstrates that~\proposed{} effectively enhances video understanding by redistributing the selection opportunities toward semantically rich tokens rather than sink tokens with limited semantics. 

\smallskip
\noindent \textbf{Results on MCQA benchmarks. }
\cref{tab:mcqa_main} presents a performance comparison across various MCQA benchmarks. From these results, we draw two observations: \textbf{1)} Existing token pruning methods show relatively moderate performance drop compared to the sharper declines seen in fine-grained video understanding task. This discrepancy suggests that MCQA benchmarks may not fully capture the limitations of token pruning, highlighting the necessity of fine-grained tasks for rigorous stress-testing. \textbf{2)}~\proposed{} consistently outperforms baseline counterparts, indicating that by explicitly guiding the pruning of sink tokens,~\proposed{} significantly bolsters the model's capacity for video understanding, even in multiple-choice format.

\begin{wraptable}{r}{0.52\textwidth}
    \centering
\vspace{-3.0ex}
\caption{Performance comparison with LLaVA-Video~\cite{zhang2024llava}.}
\setlength{\tabcolsep}{2.5pt}
\resizebox{0.93\linewidth}{!}{
\begin{tabular}{l|c|cc|cc|c}
\toprule
\multicolumn{1}{c|}{\multirow{2}{*}{\textbf{Method}}} & \multirow{2}{*}{\textbf{\begin{tabular}[c]{@{}c@{}}Retention\\ Ratio\end{tabular}}} & \multicolumn{2}{c|}{\textbf{EventHallusion}} & \multicolumn{2}{c|}{\textbf{VideoComp}}                & \multirow{2}{*}{\textbf{\begin{tabular}[c]{@{}c@{}}Performance\\ Drop Rate\end{tabular}}} \\
\multicolumn{1}{c|}{}                                 &                                                                                     & \textbf{Binary}       & \textbf{Desc.}       & \textbf{Act.}             & \textbf{YC}               &                                                                                           \\ \midrule \midrule
Vanilla                                               & 100\%                                                                               & 69.68                 & 44.37                & 71.62                     & 70.52                     &   -                                                                                        \\ \midrule
PruneVid                                              & \multirow{7}{*}{15\%}                                                               & 62.59                 & 33.11                &              66.95             &    68.72                       &  11.62\%                                                                                          \\
Holitom                                               &                                                                                     & 67.97                 & 44.37                & 71.62 & 68.21 & 1.43\%                                                                                    \\
\rowcolor{gainsboro}Holitom+\proposed{}               &      \cellcolor{white}                                                                               & 68.70                 & 45.36                & 71.26                     & 68.61                     & 0.60\%                                                                                    \\
FastVid                                               &                                                                                     & 59.17                 & 35.76                & 57.01                     & 59.55                     & 17.61\%                                                                                   \\
\rowcolor{gainsboro}FastVid+\proposed{}               &     \cellcolor{white}                                                                                 & 64.79                 & 37.75                &           66.11                &       68.81                    &  7.84\%                                                                                         \\
VisionZip                                             &                                                                                     & 59,41                 & 30.13                & 67.78                     & 66.92                     & 14.19\%                                                                                   \\
\rowcolor{gainsboro}VisionZip+\proposed{}             &       \cellcolor{white}                                                                               & 66.75                 & 42.05                &      70.42                     &   69.07                        & 3.29\%                                                                                   \\ \midrule
PruneVid                                              & \multirow{7}{*}{10\%}                                                               & 57.96                 & 31.46                & 65.51                     & 67.52                     & 14.68\%                                                                                   \\
Holitom                                               &                                                                                     & 65.77                 & 40.40                & 69.64 & 67.87 & 5.27\%                                                                                    \\
\rowcolor{gainsboro}Holitom+\proposed{}               &   \cellcolor{white}                                                                                   & 66.50                 & 45.03                & 70.78                     & 69.07                     & 1.58\%                                                                                    \\
FastVid                                               &                                                                                     & 55.35                 & 35.10                & 56.65                     & 59.04                     & 19.66\%                                                                                   \\
\rowcolor{gainsboro}FastVid+\proposed{}               &  \cellcolor{white}                                                                                    & 62.59                 & 37.42                & 65.51                     & 67.52                     & 9.66\%                                                                                    \\
VisionZip                                             &                                                                                     & 44.99                 & 21.85                & 64.07                     & 64.10                     & 26.46\%                                                                                   \\
\rowcolor{gainsboro}VisionZip+\proposed{}             &  \cellcolor{white}                                                                                    & 63.57                 & 38.74                & 69.34                     & 68.04                     & 7.04\%                                                                                    \\ \bottomrule
\end{tabular}
}
\label{tab:different_backbone}
\vspace{-4ex}
\end{wraptable}

\smallskip
\noindent \textbf{Results on Cross-Backbone. } To further validate the effectiveness of~\proposed{}, we extend our experiments to the LLaVA-Video-7B~\cite{zhang2024llava} and evaluate performance on EventHallusion and VideoComp. As shown in~\cref{tab:different_backbone}, we observe that applying~\proposed{} to various frameworks (VisionZip, FastVid, Holitom) consistently yields performance gains over their respective baselines. This result aligns with our earlier observations on LLaVA-OneVision, demonstrating the adaptability of~\proposed{}.

\vspace{-2ex}
\subsection{In-Depth Analysis}
Regarding a deeper analysis, we examine the performance of~\proposed{} at a retention ratio of 0.1 using the LLaVA-OneVision backbone.

\begin{wraptable}{r}{0.52\textwidth}
    \centering
\vspace{-8.0ex}
\caption{Ablation studies of \proposed{}.}
\resizebox{0.93\linewidth}{!}{
\begin{tabular}{c|cc|cc|cc|c}
\toprule
\multirow{2}{*}{\textbf{Method}}                                         & \multirow{2}{*}{\textbf{STSP}} & \multirow{2}{*}{\textbf{STTP}} & \multicolumn{2}{c|}{\textbf{EventHallusion}} & \multicolumn{2}{c|}{\textbf{VideoComp}} & \multirow{2}{*}{\textbf{\begin{tabular}[c]{@{}c@{}}Performance\\ Drop Rate\end{tabular}}} \\
                                                                         &                                &                                & Binary                & Desc.                & Act                & YC                 &                                                                                           \\ \midrule \midrule
Vanilla                                                                  & \textbf{}                      & \textbf{}                      & 63.33                 & 38.74                & 70.06              & 70.95              & -                                                                                         \\ \midrule
\multirow{2}{*}{\begin{tabular}[c]{@{}c@{}}VisionZip\\ (S)\end{tabular}} &                                &                                & 50.37                 & 28.15                & 65.39              & 64.61              & 20.46\%                                                                                   \\
                                                                         & \ding{51}                      &                                & \textbf{60.39}        & \textbf{36.09}       & \textbf{67.78}     & \textbf{68.89}     & \textbf{4.64\%}                                                                           \\ \midrule
\multirow{3}{*}{\begin{tabular}[c]{@{}c@{}}Holitom\\ (T+S)\end{tabular}} &                                &                                & 60.88                 & 36.75                & 68.38              & 68.98              & 3.87\%                                                                                    \\
                                                                         & \ding{51}                      &                                & 62.10                 & 37.09                & 68.02              & \textbf{69.24}     & 1.94\%                                                                                    \\
                                                                         & \ding{51}                      & \ding{51}                      & \textbf{62.59}        & \textbf{39.74}       & \textbf{69.24}     & 69.10              & \textbf{1.17\%}                                                                           \\ \bottomrule
\end{tabular}
}
\label{tab:ablation}
\vspace{-4ex}
\end{wraptable}

\smallskip
\noindent \textbf{Ablation Studies. }In~\cref{tab:ablation}, we conduct ablation studies to understand the effect of each component. To this end, we apply~\proposed{} on VisionZip for spatial pruning (S) and Holitom for temporal+spatial pruning (T+S) and evaluate the performance on EventHallusion and VideoComp datasets. We observe that the STSP module significantly improves the performance of VisionZip. It indicates that the STSP module effectively guides the pruning of sink tokens while selecting truly salient tokens. Furthermore, even if Holitom implicitly suppresses sink tokens via temporal pruning, discussed in~\cref{sec:sink_token_obstacle}, thereby enhancing video understanding, the STSP modules further improve performance, indicating that there is room for improvement to facilitate the pruning of sink tokens. Finally, incorporating the STTP module into Holitom achieves the best overall performance, demonstrating that the STTP module effectively complements spatial pruning. These results validate the effectiveness of both the STSP and STTP modules.

\begin{wrapfigure}{r}{0.52\linewidth}
  \centering
  \vspace{-28pt} 
  \includegraphics[width=0.99\linewidth]{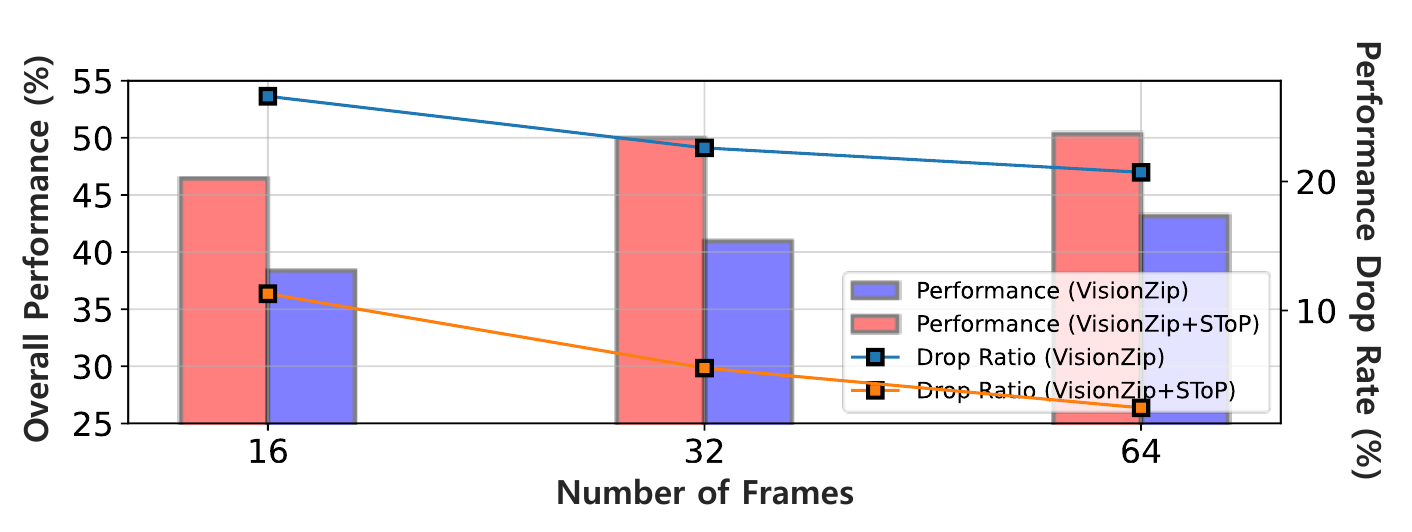}
  \vspace{-5ex}
  \caption{Performance over the different number of frames.}
  \label{fig:frame}
  \vspace{-22pt}
\end{wrapfigure}

\smallskip
\noindent \textbf{Different Number of Frames. }
To investigate how performance varies with the different number of frames (16, 32, 64), we evaluate VisionZip and VisionZip+\proposed{} on the EventHallusion dataset. For each setting, we report the overall performance and the performance drop rate compared to the corresponding vanilla model. As illustrated in~\cref{fig:frame}, we have the following observations. \textbf{1) }The performance of VisionZip linearly increases as the number of frames increases. This suggests that benefit of incorporating additional visual cues outweighs the semantic dilution potentially caused by the accumulation of sink tokens in larger frames. However, this gain comes at the cost of increased inference latency, posing a challenge for real-time applications. \textbf{2) }\proposed{} consistently yields substantial gains over the VisionZip baseline across all configurations. Especially, we observe that~\proposed{} with only 16 frames surpasses the baseline with 64 frames, highlighting the superiority of~\proposed{} in scenarios with a strict token budget. By enhancing the models' fundamental video understanding,~\proposed{} allows for high-accuracy inference using significantly fewer frames, thereby optimizing the trade-off between performance and computational efficiency.

\begin{wrapfigure}{r}{0.48\linewidth}
  \centering
  \vspace{-26pt} 
  \includegraphics[width=0.95\linewidth]{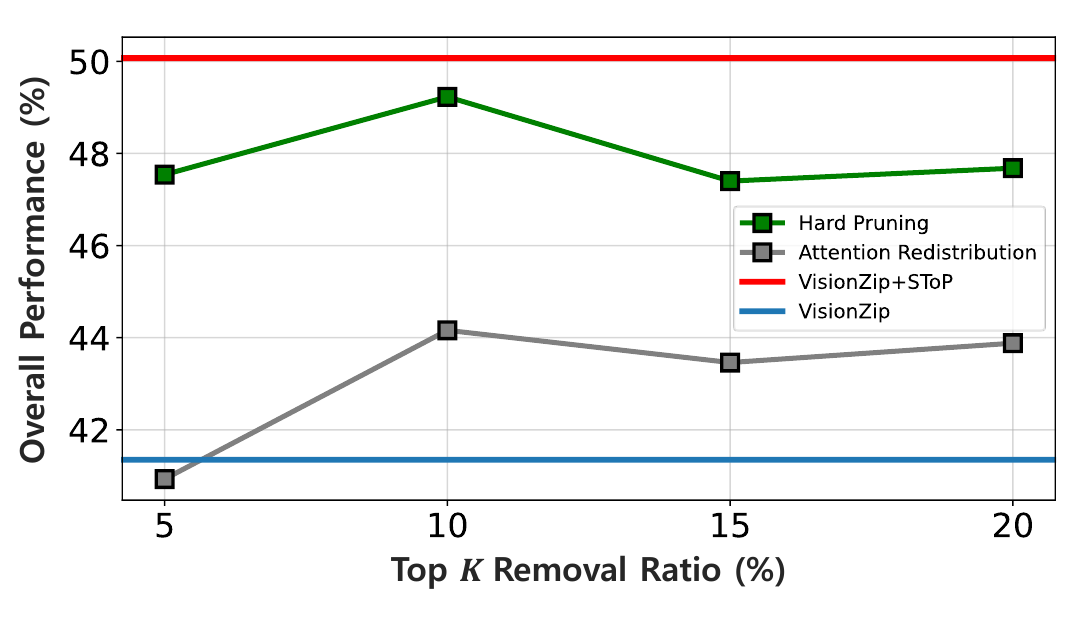}
  \vspace{-3ex}
  \caption{Performance with naive approaches.}
  \label{fig:naive_approach_exp}
  \vspace{-22pt}
\end{wrapfigure}
\smallskip
\noindent \textbf{Comparison with Naive Approaches. }To better understand the underlying mechanics of~\proposed{} beyond the main results, we compare it against two naive approaches applied to VisionZip on the EventHallusion dataset. For these approaches, we first identify tokens with the top $K$\% highest attention weights per frame, which often behave as sink tokens (see~\cref{sec:source_high_freq}). We then apply two distinct approaches:  \textit{Hard Pruning}: these top $K$\% tokens are deliberately discarded, after which we apply VisionZip, where tokens with high attention are selected. \textit{Attention Redistribution}: Since mitigation of sink tokens was explored by prior work, we attempt to adopt previous work for visual token pruning to see if it is effective. Therefore, following~\cite{kang2025see}, we redistribute the attention weights from these $K$ tokens to the remaining tokens before selection.~\cref{fig:naive_approach_exp} illustrates the performance across various $K$ (5\% to 20\%), from which we draw the following observations: \textbf{1) }\textit{Hard Pruning} consistently outperforms the original VisionZip. This confirms our hypothesis that the most attended tokens are often semantically sparse sink tokens. From the result of peaking at $K=10\%$, we hypothesize that they are likely to stay around tokens with top 10\% highest attention. \textbf{2) }\textit{Attention Redistribution} underperforms compared to both \textit{Hard pruning} and~\proposed{}. This indicates that visual token pruning requires a different mechanism in that sink-token mitigation methods mainly focus on adjusting inherent information flow, whereas for visual token pruning, explicitly guiding the pruning of sink tokens is necessary. In this vein, \textit{Hard pruning} that directly decides to prune sink tokens is more effective than this approach. \textbf{3) }\proposed{} achieves the highest performance, outperforming all naive approaches. While the hard pruning approach risks discarding truly salient tokens that happen to have high attention, ~\proposed{} combines attention-based information density with a sink score to better distinguish truly salient tokens from sink tokens, leading to more reliable selection of informative tokens.

\begin{wraptable}{r}{0.6\textwidth}
    \centering
\vspace{-8.0ex}
\caption{Comparison with feature-based spatial pruning approach.}
\setlength{\tabcolsep}{2.5pt}
\resizebox{0.93\linewidth}{!}{
\begin{tabular}{l|c|cc|cc|c}
\toprule
\multirow{2}{*}{\textbf{Method}} & \multirow{2}{*}{\textbf{Retention}} & \multicolumn{2}{c|}{\textbf{EventHallusion}} & \multicolumn{2}{c|}{\textbf{VideoComp}} & \multirow{2}{*}{\textbf{\begin{tabular}[c]{@{}c@{}}Performance\\ Drop Rate\end{tabular}}} \\
                                 &                                     & Binary                & Desc.                & Act.               & YC                 &                                                                                           \\ \midrule \midrule
Vanilla                          & 100\%                               & 63.33                 & 38.74                & 70.06              & 70.95              & -                                                                                         \\ \midrule
Feature-based Pruning  & \multirow{3}{*}{15\%}               & 57.95                 & 33.44                & \textbf{68.98}     & 69.15              & 6.56\%                                                                                      \\
Attention-based Pruning    &                                     & 58.44                 & 34.44                & 67.78              & 68.29              & 6.46\%                                                                                      \\
+\proposed{}                     &                                     & \textbf{61.61}        & \textbf{38.08}       & 68.38              & \textbf{69.84}     & \textbf{2.10\%}                                                                             \\ \midrule
Feature-based Pruning  & \multirow{3}{*}{10\%}               & 55.99                 & 35.43                & \textbf{68.14}     & 68.72              & 6.50\%                                                                                      \\
Attention-based Pruning   &                                     & 50.37                 & 28.15                & 65.39              & 64.61              & 15.85\%                                                                                     \\
+\proposed{}                     &                                     & \textbf{60.39}        & \textbf{36.09}       & 67.78              & \textbf{68.89}     & \textbf{4.41\%}                                                                             \\ \bottomrule
\end{tabular}
}
\label{tab:density}
\vspace{-4ex}
\end{wraptable}

\vspace{-1ex}
\subsection{Comparison with Other Pruning Approaches}
\label{sec:diversity}
To explore the effectiveness of~\proposed{} against other token pruning approach, we compare it against feature-based spatial pruning, a widely adopted alternative that may potentially avoids the sink token issue by not relying on attention scores. Specifically, we employ Density Peaks Clustering with $k$-Nearest Neighbors (DPC-KNN~\cite{du2016study}) algorithm per frame for spatial pruning, following established frameworks in recent work~\cite{dhouib2025pact,huang2025prunevid,ma2025mmg,shen2025fastvid}. For the attention-based baseline, we use VisionZip. As shown in~\cref{tab:density}, we have the following observations: \textbf{1) } At a low retention ratio of 10\%, the standard attention-based pruning suffers a significant performance drop. On the other hand, feature-based pruning shows higher performance, as it bypasses the distractions caused by sink tokens. \textbf{2) } Applying~\proposed{} to the attention-based pruning yields substantial improvements, eventually surpassing the feature-based pruning. This indicates that while naive attention-based pruning is hindered by sink tokens,~\proposed{} effectively mitigates this bottleneck, validating its robustness under aggressive spatial pruning.


\vspace{-1ex}
\section{Related Works}
\vspace{-1ex}
\label{sec:related_work}
\noindent \textbf{Visual Token pruning in Video LLMs. }
Recent progress in Video LLMs have introduced visual token pruning to mitigate the high computational overhead associated with processing multiple frames.
To achieve this, some methods reduce visual tokens \textit{inside the LLM} to optimize KV cache memory (e.g., FastV~\cite{chen2024image}, DSTP~\cite{kim2026and}). In contrast, we focus on pruning \textit{before} the visual tokens are fed into the LLM, which maintains full compatibility with optimized attention kernels (e.g., FlashAttention~\cite{dao2022flashattention}). From this perspective, prior work can be broadly categorized into spatial-only pruning and temporal+spatial pruning approach. For \textbf{spatial-only} pruning, VisionZip~\cite{yang2025visionzip} and FasterVLM~\cite{zhang2024cls} utilize vision encoder's attention weights to select salient tokens. FastVid~\cite{shen2025fastvid} adopts a similar attention-based selection strategy, and further applies density-based token merging to the remaining tokens. CDPruner~\cite{zhang2025beyond} reframes pruning through a detrimental point process to enable dynamic token pruning. 
To further exploit redundancy across frames, several methods perform \textbf{temporal+spatial pruning}.
PruneVid~\cite{huang2025prunevid} and HoliTom~\cite{shao2025holitom} segment videos into clips to first reduce temporal redundancy via inter-frame similarity, followed by spatial clustering (e.g., DPC-KNN~\cite{du2016study}) or attention-based pruning. More recently, STTM~\cite{hyun2025multi} and FlashVid~\cite{fan2026flashvid} have introduced tree-based structures for hierarchical spatio-temporal reduction. In contrast, Dycoke~\cite{tao2025dycoke} performs only temporal pruning, which limits the achievable token reduction (up to 25\%). 
Despite their efficiency, existing works are predominantly evaluated on coarse-grained video understanding tasks, such as MCQA. On the other hand, we shifts the focus toward fine-grained video understanding (e.g., hallucination or compositional reasoning), where we observe that sink token is a key obstacle, especially at low token retention rates.

\smallskip
\noindent \textbf{Attention Sink.}
The attention sink phenomenon in the transformer-based models refers to the tendency for a small subset of low-semantic tokens to attract a disproportionate amount of attention. This phenomenon was first documented in Large Language Models (LLMs), where certain tokens with minimal semantic meaning (e.g., BOS, SEP token) consistently receive extremely high attention scores~\cite{xiao2023efficient,sun2024massive}. Similar to the sinks found in LLMs, 
several works~\cite{jiang2025vision,feng2026edit,kim2025activation,lu2025artifacts,sinkssink,darcet2023vision} have identified an analogous phenomenon in Vision Transformers (ViTs), where the small subset of visual tokens with low semantics receive excessive attention. While prior research has sought to mitigate these sinks to improve ViT interpretability of ViTs~\cite{darcet2023vision,jiang2025vision} or strengthen internal information flow~\cite{feng2026edit,lu2025artifacts,sinkssink}, their impact on visual token pruning remains underexplored. In this work, we demonstrate that these sink tokens are a key obstacle to effective visual token pruning in fine-grained video tasks, motivating us to propose~\proposed{}.

\vspace{-1.5ex}
\section{Conclusion}
\vspace{-1ex}
In this work, we revisit a key assumption in visual token pruning: whether existing methods indeed preserve fine-grained visual cues.
We argue that this has been under-examined since most prior work is validated mainly on MCQA benchmarks, where coarse-grained cues are often sufficient. From our analysis, we identify sink tokens as a key bottleneck for fine-grained video understanding. Building on this insight, we propose~\proposed{}, which introduces a sink score to quantify each token's tendency to behave as a sink. We then use this score within established spatial and temporal pruning frameworks, implemented via the STSP and STTP modules, respectively. Extensive experiments show that~\proposed{} consistently strengthens fine-grained video understanding under aggressive pruning, delivering clear performance gains, especially at a 10\% retention ratio. 
Moreover,~\proposed{} makes few-frame inference substantially more effective, outperforming the baseline that requires many more frames to achieve comparable understanding. Regarding limitations and future work, please refer to Appendix~\ref{app_sec:limitation}.

More broadly, we believe this work offers two key contributions to the visual token pruning community: 1) \textit{Spatial Insight}: explicitly addressing sink tokens is crucial for capturing the fine-grained visual cues in attention-based pruning. 2) \textit{Temporal Insight}: temporal pruning can serve as a potent sink token suppressor, extending its utility beyond mere temporal redundancy reduction.

\clearpage  


%
%
\bibliographystyle{splncs04}
\bibliography{main}

@article{wang2024cls,
  title={[CLS] token tells everything needed for training-free efficient MLLMs},
  author={Wang, Ao and Sun, Fengyuan and Chen, Hui and Lin, Zijia and Han, Jungong and Ding, Guiguang},
  journal={arXiv preprint arXiv:2412.05819},
  year={2024}
}

@article{zhang2025videollama,
  title={Videollama 3: Frontier multimodal foundation models for image and video understanding},
  author={Zhang, Boqiang and Li, Kehan and Cheng, Zesen and Hu, Zhiqiang and Yuan, Yuqian and Chen, Guanzheng and Leng, Sicong and Jiang, Yuming and Zhang, Hang and Li, Xin and others},
  journal={arXiv preprint arXiv:2501.13106},
  year={2025}
}

@article{li2024llava,
  title={Llava-onevision: Easy visual task transfer},
  author={Li, Bo and Zhang, Yuanhan and Guo, Dong and Zhang, Renrui and Li, Feng and Zhang, Hao and Zhang, Kaichen and Zhang, Peiyuan and Li, Yanwei and Liu, Ziwei and others},
  journal={arXiv preprint arXiv:2408.03326},
  year={2024}
}

@inproceedings{lin2024video,
  title={Video-llava: Learning united visual representation by alignment before projection},
  author={Lin, Bin and Ye, Yang and Zhu, Bin and Cui, Jiaxi and Ning, Munan and Jin, Peng and Yuan, Li},
  booktitle={Proceedings of the 2024 conference on empirical methods in natural language processing},
  pages={5971--5984},
  year={2024}
}

@article{bai2023qwen,
  title={Qwen technical report},
  author={Bai, Jinze and Bai, Shuai and Chu, Yunfei and Cui, Zeyu and Dang, Kai and Deng, Xiaodong and Fan, Yang and Ge, Wenbin and Han, Yu and Huang, Fei and others},
  journal={arXiv preprint arXiv:2309.16609},
  year={2023}
}

@inproceedings{li2024llama,
  title={Llama-vid: An image is worth 2 tokens in large language models},
  author={Li, Yanwei and Wang, Chengyao and Jia, Jiaya},
  booktitle={European Conference on Computer Vision},
  pages={323--340},
  year={2024},
  organization={Springer}
}

@inproceedings{maaz2024video,
  title={Video-chatgpt: Towards detailed video understanding via large vision and language models},
  author={Maaz, Muhammad and Rasheed, Hanoona and Khan, Salman and Khan, Fahad},
  booktitle={Proceedings of the 62nd Annual Meeting of the Association for Computational Linguistics (Volume 1: Long Papers)},
  pages={12585--12602},
  year={2024}
}

@article{xu2024pllava,
  title={Pllava: Parameter-free llava extension from images to videos for video dense captioning},
  author={Xu, Lin and Zhao, Yilin and Zhou, Daquan and Lin, Zhijie and Ng, See Kiong and Feng, Jiashi},
  journal={arXiv preprint arXiv:2404.16994},
  year={2024}
}

@article{shen2024longvu,
  title={Longvu: Spatiotemporal adaptive compression for long video-language understanding},
  author={Shen, Xiaoqian and Xiong, Yunyang and Zhao, Changsheng and Wu, Lemeng and Chen, Jun and Zhu, Chenchen and Liu, Zechun and Xiao, Fanyi and Varadarajan, Balakrishnan and Bordes, Florian and others},
  journal={arXiv preprint arXiv:2410.17434},
  year={2024}
}

@inproceedings{yang2025visionzip,
  title={Visionzip: Longer is better but not necessary in vision language models},
  author={Yang, Senqiao and Chen, Yukang and Tian, Zhuotao and Wang, Chengyao and Li, Jingyao and Yu, Bei and Jia, Jiaya},
  booktitle={Proceedings of the IEEE/CVF Conference on Computer Vision and Pattern Recognition},
  pages={19792--19802},
  year={2025}
}

@inproceedings{huang2025prunevid,
  title={Prunevid: Visual token pruning for efficient video large language models},
  author={Huang, Xiaohu and Zhou, Hao and Han, Kai},
  booktitle={Findings of the Association for Computational Linguistics: ACL 2025},
  pages={19959--19973},
  year={2025}
}

@article{shen2025fastvid,
  title={Fastvid: Dynamic density pruning for fast video large language models},
  author={Shen, Leqi and Gong, Guoqiang and He, Tao and Zhang, Yifeng and Liu, Pengzhang and Zhao, Sicheng and Ding, Guiguang},
  journal={arXiv preprint arXiv:2503.11187},
  year={2025}
}

@article{zhang2025beyond,
  title={Beyond attention or similarity: Maximizing conditional diversity for token pruning in mllms},
  author={Zhang, Qizhe and Liu, Mengzhen and Li, Lichen and Lu, Ming and Zhang, Yuan and Pan, Junwen and She, Qi and Zhang, Shanghang},
  journal={arXiv preprint arXiv:2506.10967},
  year={2025}
}

@article{zhang2024eventhallusion,
  title={Eventhallusion: Diagnosing event hallucinations in video llms},
  author={Zhang, Jiacheng and Jiao, Yang and Chen, Shaoxiang and Zhao, Na and Tan, Zhiyu and Li, Hao and Ma, Xingjun and Chen, Jingjing},
  journal={arXiv preprint arXiv:2409.16597},
  year={2024}
}

@article{shao2025holitom,
  title={Holitom: Holistic token merging for fast video large language models},
  author={Shao, Kele and Tao, Keda and Qin, Can and You, Haoxuan and Sui, Yang and Wang, Huan},
  journal={arXiv preprint arXiv:2505.21334},
  year={2025}
}

@inproceedings{li2024mvbench,
  title={Mvbench: A comprehensive multi-modal video understanding benchmark},
  author={Li, Kunchang and Wang, Yali and He, Yinan and Li, Yizhuo and Wang, Yi and Liu, Yi and Wang, Zun and Xu, Jilan and Chen, Guo and Luo, Ping and others},
  booktitle={Proceedings of the IEEE/CVF Conference on Computer Vision and Pattern Recognition},
  pages={22195--22206},
  year={2024}
}

@article{mangalam2023egoschema,
  title={Egoschema: A diagnostic benchmark for very long-form video language understanding},
  author={Mangalam, Karttikeya and Akshulakov, Raiymbek and Malik, Jitendra},
  journal={Advances in Neural Information Processing Systems},
  volume={36},
  pages={46212--46244},
  year={2023}
}

@inproceedings{fu2025video,
  title={Video-mme: The first-ever comprehensive evaluation benchmark of multi-modal llms in video analysis},
  author={Fu, Chaoyou and Dai, Yuhan and Luo, Yongdong and Li, Lei and Ren, Shuhuai and Zhang, Renrui and Wang, Zihan and Zhou, Chenyu and Shen, Yunhang and Zhang, Mengdan and others},
  booktitle={Proceedings of the IEEE/CVF conference on computer vision and pattern recognition},
  pages={24108--24118},
  year={2025}
}

@article{fan2026flashvid,
  title={FlashVID: Efficient Video Large Language Models via Training-free Tree-based Spatiotemporal Token Merging},
  author={Fan, Ziyang and Chen, Keyu and Xing, Ruilong and Li, Yulin and Jiang, Li and Tian, Zhuotao},
  journal={arXiv preprint arXiv:2602.08024},
  year={2026}
}

@article{xiao2023efficient,
  title={Efficient streaming language models with attention sinks},
  author={Xiao, Guangxuan and Tian, Yuandong and Chen, Beidi and Han, Song and Lewis, Mike},
  journal={arXiv preprint arXiv:2309.17453},
  year={2023}
}

@article{jiang2025vision,
  title={Vision Transformers Don't Need Trained Registers},
  author={Jiang, Nick and Dravid, Amil and Efros, Alexei and Gandelsman, Yossi},
  journal={arXiv preprint arXiv:2506.08010},
  year={2025}
}

@inproceedings{zhai2023sigmoid,
  title={Sigmoid loss for language image pre-training},
  author={Zhai, Xiaohua and Mustafa, Basil and Kolesnikov, Alexander and Beyer, Lucas},
  booktitle={Proceedings of the IEEE/CVF international conference on computer vision},
  pages={11975--11986},
  year={2023}
}

@inproceedings{lin2024vila,
  title={Vila: On pre-training for visual language models},
  author={Lin, Ji and Yin, Hongxu and Ping, Wei and Molchanov, Pavlo and Shoeybi, Mohammad and Han, Song},
  booktitle={Proceedings of the IEEE/CVF conference on computer vision and pattern recognition},
  pages={26689--26699},
  year={2024}
}

@inproceedings{shang2025llava,
  title={Llava-prumerge: Adaptive token reduction for efficient large multimodal models},
  author={Shang, Yuzhang and Cai, Mu and Xu, Bingxin and Lee, Yong Jae and Yan, Yan},
  booktitle={Proceedings of the IEEE/CVF International Conference on Computer Vision},
  pages={22857--22867},
  year={2025}
}

@article{zhang2024cls,
  title={[CLS] Attention is All You Need for Training-Free Visual Token Pruning: Make VLM Inference Faster},
  author={Zhang, Qizhe and Cheng, Aosong and Lu, Ming and Zhuo, Zhiyong and Wang, Minqi and Cao, Jiajun and Guo, Shaobo and She, Qi and Zhang, Shanghang},
  journal={arXiv e-prints},
  pages={arXiv--2412},
  year={2024}
}

@inproceedings{hyun2025multi,
  title={Multi-granular spatio-temporal token merging for training-free acceleration of video llms},
  author={Hyun, Jeongseok and Hwang, Sukjun and Han, Su Ho and Kim, Taeoh and Lee, Inwoong and Wee, Dongyoon and Lee, Joon-Young and Kim, Seon Joo and Shim, Minho},
  booktitle={Proceedings of the IEEE/CVF International Conference on Computer Vision},
  pages={23990--24000},
  year={2025}
}

@inproceedings{tao2025dycoke,
  title={Dycoke: Dynamic compression of tokens for fast video large language models},
  author={Tao, Keda and Qin, Can and You, Haoxuan and Sui, Yang and Wang, Huan},
  booktitle={Proceedings of the Computer Vision and Pattern Recognition Conference},
  pages={18992--19001},
  year={2025}
}

@article{tatler2007central,
  title={The central fixation bias in scene viewing: Selecting an optimal viewing position independently of motor biases and image feature distributions},
  author={Tatler, Benjamin W},
  journal={Journal of vision},
  volume={7},
  number={14},
  pages={4--4},
  year={2007},
  publisher={The Association for Research in Vision and Ophthalmology}
}

@inproceedings{chen2024image,
  title={An image is worth 1/2 tokens after layer 2: Plug-and-play inference acceleration for large vision-language models},
  author={Chen, Liang and Zhao, Haozhe and Liu, Tianyu and Bai, Shuai and Lin, Junyang and Zhou, Chang and Chang, Baobao},
  booktitle={European Conference on Computer Vision},
  pages={19--35},
  year={2024},
  organization={Springer}
}

@article{dao2022flashattention,
  title={Flashattention: Fast and memory-efficient exact attention with io-awareness},
  author={Dao, Tri and Fu, Dan and Ermon, Stefano and Rudra, Atri and R{\'e}, Christopher},
  journal={Advances in neural information processing systems},
  volume={35},
  pages={16344--16359},
  year={2022}
}

@article{du2016study,
  title={Study on density peaks clustering based on k-nearest neighbors and principal component analysis},
  author={Du, Mingjing and Ding, Shifei and Jia, Hongjie},
  journal={Knowledge-Based Systems},
  volume={99},
  pages={135--145},
  year={2016},
  publisher={Elsevier}
}

@article{kim2025activation,
  title={Activation Quantization of Vision Encoders Needs Prefixing Registers},
  author={Kim, Seunghyeon and Kim, Jinho and Yeom, Taesun and Park, Wonpyo and Kim, Kyuyeun and Lee, Jaeho},
  journal={arXiv preprint arXiv:2510.04547},
  year={2025}
}

@article{lu2025artifacts,
  title={Artifacts and attention sinks: Structured approximations for efficient vision transformers},
  author={Lu, Andrew and Liao, Wentinn and Wang, Liuhui and Yang, Huzheng and Shi, Jianbo},
  journal={arXiv preprint arXiv:2507.16018},
  year={2025}
}

@inproceedings{feng2026edit,
  title={EDIT: enhancing vision transformers by mitigating attention sink through an encoder-decoder architecture},
  author={Feng, Wenfeng and Wang, Hongxiang and Wang, Jianlong and Zhang, Xin and Zhao, Jingjing and Liang, Yueyue and Chen, Xiang and Han, Duokui},
  booktitle={International Conference on Optoelectronics, Computer Science, and Algorithms (OCSA 2025)},
  volume={14008},
  pages={246--259},
  year={2026},
  organization={SPIE}
}

@article{sinkssink,
  title={To Sink or Not to Sink: Visual Information Pathways in Large Vision-Language Models},
  author={Sinks, E All VLM and Sinks, D LLM-emerged and Sinks, C Propagated ViT and Sinks, BVIT and Image, A Given}
}

@article{sun2024massive,
  title={Massive activations in large language models},
  author={Sun, Mingjie and Chen, Xinlei and Kolter, J Zico and Liu, Zhuang},
  journal={arXiv preprint arXiv:2402.17762},
  year={2024}
}

@article{darcet2023vision,
  title={Vision transformers need registers},
  author={Darcet, Timoth{\'e}e and Oquab, Maxime and Mairal, Julien and Bojanowski, Piotr},
  journal={arXiv preprint arXiv:2309.16588},
  year={2023}
}

@inproceedings{kim2025videocomp,
  title={Videocomp: Advancing fine-grained compositional and temporal alignment in video-text models},
  author={Kim, Dahun and Piergiovanni, AJ and Mallya, Ganesh and Angelova, Anelia},
  booktitle={Proceedings of the IEEE/CVF Conference on Computer Vision and Pattern Recognition},
  pages={29060--29070},
  year={2025}
}

@inproceedings{xiao2021next,
  title={Next-qa: Next phase of question-answering to explaining temporal actions},
  author={Xiao, Junbin and Shang, Xindi and Yao, Angela and Chua, Tat-Seng},
  booktitle={Proceedings of the IEEE/CVF conference on computer vision and pattern recognition},
  pages={9777--9786},
  year={2021}
}

@article{wu2407longvideobench,
  title={Longvideobench: A benchmark for long-context interleaved video-language understanding, 2024},
  author={Wu, Haoning and Li, Dongxu and Chen, Bei and Li, Junnan},
  journal={URL https://arxiv. org/abs/2407},
  volume={15754},
  number={8}
}

@inproceedings{zhou2025mlvu,
  title={Mlvu: Benchmarking multi-task long video understanding},
  author={Zhou, Junjie and Shu, Yan and Zhao, Bo and Wu, Boya and Liang, Zhengyang and Xiao, Shitao and Qin, Minghao and Yang, Xi and Xiong, Yongping and Zhang, Bo and others},
  booktitle={Proceedings of the IEEE/CVF Conference on Computer Vision and Pattern Recognition},
  pages={13691--13701},
  year={2025}
}

@article{zhang2024llava,
  title={Llava-video: Video instruction tuning with synthetic data},
  author={Zhang, Yuanhan and Wu, Jinming and Li, Wei and Li, Bo and Ma, Zejun and Liu, Ziwei and Li, Chunyuan},
  journal={arXiv preprint arXiv:2410.02713},
  year={2024}
}

@article{singh2025openai,
  title={Openai gpt-5 system card},
  author={Singh, Aaditya and Fry, Adam and Perelman, Adam and Tart, Adam and Ganesh, Adi and El-Kishky, Ahmed and McLaughlin, Aidan and Low, Aiden and Ostrow, AJ and Ananthram, Akhila and others},
  journal={arXiv preprint arXiv:2601.03267},
  year={2025}
}

@article{chen2024cg,
  title={Cg-bench: Clue-grounded question answering benchmark for long video understanding},
  author={Chen, Guo and Liu, Yicheng and Huang, Yifei and He, Yuping and Pei, Baoqi and Xu, Jilan and Wang, Yali and Lu, Tong and Wang, Limin},
  journal={arXiv preprint arXiv:2412.12075},
  year={2024}
}

@article{kang2025see,
  title={See what you are told: Visual attention sink in large multimodal models},
  author={Kang, Seil and Kim, Jinyeong and Kim, Junhyeok and Hwang, Seong Jae},
  journal={arXiv preprint arXiv:2503.03321},
  year={2025}
}

@inproceedings{dhouib2025pact,
  title={Pact: Pruning and clustering-based token reduction for faster visual language models},
  author={Dhouib, Mohamed and Buscaldi, Davide and Vanier, Sonia and Shabou, Aymen},
  booktitle={Proceedings of the Computer Vision and Pattern Recognition Conference},
  pages={14582--14592},
  year={2025}
}

@article{ma2025mmg,
  title={Mmg-vid: Maximizing marginal gains at segment-level and token-level for efficient video llms},
  author={Ma, Junpeng and Zhang, Qizhe and Lu, Ming and Wang, Zhibin and Zhou, Qiang and Song, Jun and Zhang, Shanghang},
  journal={arXiv preprint arXiv:2508.21044},
  year={2025}
}

@inproceedings{qiu2025step,
  title={STEP: Enhancing Video-LLMs' Compositional Reasoning by Spatio-Temporal Graph-guided Self-Training},
  author={Qiu, Haiyi and Gao, Minghe and Qian, Long and Pan, Kaihang and Yu, Qifan and Li, Juncheng and Wang, Wenjie and Tang, Siliang and Zhuang, Yueting and Chua, Tat-Seng},
  booktitle={Proceedings of the IEEE/CVF Conference on Computer Vision and Pattern Recognition},
  pages={3284--3294},
  year={2025}
}

@article{grattafiori2024llama,
  title={The llama 3 herd of models},
  author={Grattafiori, Aaron and Dubey, Abhimanyu and Jauhri, Abhinav and Pandey, Abhinav and Kadian, Abhishek and Al-Dahle, Ahmad and Letman, Aiesha and Mathur, Akhil and Schelten, Alan and Vaughan, Alex and others},
  journal={arXiv preprint arXiv:2407.21783},
  year={2024}
}

@inproceedings{alvar2025divprune,
  title={Divprune: Diversity-based visual token pruning for large multimodal models},
  author={Alvar, Saeed Ranjbar and Singh, Gursimran and Akbari, Mohammad and Zhang, Yong},
  booktitle={Proceedings of the Computer Vision and Pattern Recognition Conference},
  pages={9392--9401},
  year={2025}
}

@inproceedings{rawal2025argus,
  title={Argus: Hallucination and omission evaluation in video-llms},
  author={Rawal, Ruchit and Shirkavand, Reza and Huang, Heng and Somepalli, Gowthami and Goldstein, Tom},
  booktitle={Proceedings of the IEEE/CVF International Conference on Computer Vision},
  pages={20280--20290},
  year={2025}
}

@misc{bai2025qwen25vltechnicalreport,
      title={Qwen2.5-VL Technical Report}, 
      author={Shuai Bai and Keqin Chen and Xuejing Liu and Jialin Wang and Wenbin Ge and Sibo Song and Kai Dang and Peng Wang and Shijie Wang and Jun Tang and Humen Zhong and Yuanzhi Zhu and Mingkun Yang and Zhaohai Li and Jianqiang Wan and Pengfei Wang and Wei Ding and Zheren Fu and Yiheng Xu and Jiabo Ye and Xi Zhang and Tianbao Xie and Zesen Cheng and Hang Zhang and Zhibo Yang and Haiyang Xu and Junyang Lin},
      year={2025},
      eprint={2502.13923},
      archivePrefix={arXiv},
      primaryClass={cs.CV},
      url={https://arxiv.org/abs/2502.13923}, 
}

@article{kim2026and,
  title={Why and When Visual Token Pruning Fails? A Study on Relevant Visual Information Shift in MLLMs Decoding},
  author={Kim, Jiwan and Kim, Kibum and Kim, Wonjoong and Lee, Byung-Kwan and Park, Chanyoung},
  journal={arXiv preprint arXiv:2604.12358},
  year={2026}
}

@article{kim2025compodistill,
  title={Compodistill: Attention distillation for compositional reasoning in multimodal llms},
  author={Kim, Jiwan and Kim, Kibum and Seo, Sangwoo and Park, Chanyoung},
  journal={arXiv preprint arXiv:2510.12184},
  year={2025}
}

\clearpage

\appendix
\newpage
\startcontents[appendix]

\newpage
\begin{center}
    \huge{\emph{Supplementary Material}}
\end{center}

\begin{center}
    \large{\emph{- Sink-Token-Aware Pruning for Fine-Grained Video Understanding in Efficient Video LLMs -}}
\end{center}
\vspace{-4ex}

\vspace*{3mm}
\rule[0pt]{\columnwidth}{1pt}
\appendix


\vspace*{.5in}



\vspace{-6ex}
\section{Difference of Sink Tokens in LLMs vs. Vision Encoders}
\vspace{-1ex}
\label{app_sec:sink_diff}
In this section, we clarify the conceptual similarities and key differences between sink tokens in LLMs and those we identify within vision encoders.

\smallskip
\noindent \textbf{Sink Tokens in LLMs.} In autoregressive LLMs~\cite{grattafiori2024llama,bai2023qwen}, sink tokens typically reside at fixed positions (e.g., the BOS token) and act as global drains that absorb disproportionate attention during inference. Previous studies~\cite{xiao2023efficient,sun2024massive} demonstrate that removing these tokens causes catastrophic performance degradation, as they play a structural role in stabilizing the softmax distribution. Thus, despite carrying minimal semantic information, sink tokens in LLMs are functionally necessary and must be preserved.

\smallskip
\noindent \textbf{Sink Tokens in Vision Encoders.} On the other hand, the sink tokens we identify in vision encoders are fundamentally different in their role. As discussed in Sec.~3.2, these tokens exhibit spatially persistent high attention scores across frames, yet they tend to correspond to semantically sparse background regions. Unlike their LLM counterparts, these tokens do not serve any structural function—when they survive pruning, they actively impair fine-grained video understanding by consuming the limited token budget at the expense of truly salient tokens.

In summary, sink tokens in both LLMs and vision encoders have two properties in common: abnormally high attention scores and low semantic content. However, their roles diverge fundamentally. Sink tokens in LLMs must be preserved to maintain softmax stability, whereas sink tokens in vision encoders should be actively pruned, as retaining them crowds out salient tokens and hinders fine-grained understanding. This contrast is the core motivation of \proposed{}, which explicitly identifies and suppresses sink tokens during pruning—a direction conceptually opposite to sink-preserving methods in the LLM literature.

\begin{wrapfigure}{r}{0.33\linewidth}
  \centering
  \vspace{-26pt} 
  \includegraphics[width=0.99\linewidth]{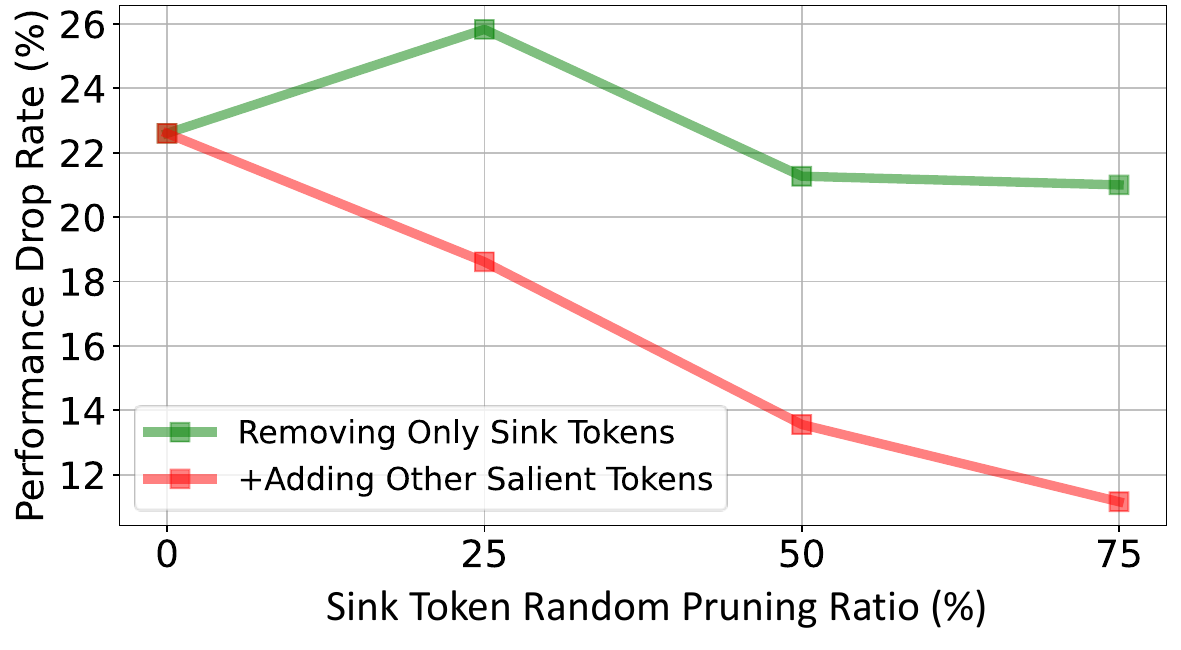}
  \vspace{-5ex}
  \caption{Performance degradation rate across varying sink token removal ratios.}
  \label{fig:naive_approach_no_add}
  \vspace{-21pt}
\end{wrapfigure}

\vspace{-1ex}
\section{Detailed Ablation of Impact of Sink Tokens}
\vspace{-1ex}
\label{app_sec:ablation_sink_token}
In Sec.~3.3 of the main paper, we demonstrated that removing sink tokens from the selected token set—while replacing them with tokens of the next highest attention scores—reduces hallucination. Here, we provide a more systematic ablation to further isolate the inherent impact of sink tokens on fine-grained video understanding.

\smallskip
\noindent \textbf{Setup. }
To directly assess the impact of sink tokens themselves, we remove them from the selected token set \textit{without} replacing them with additional tokens, thus operating under a strictly reduced token budget. Herein, we define sink tokens as the top 5\% highest-frequency tokens in Fig.~2(b)\footnote{We use top 5\% instead of top 10\% (Sec.~3.3), as removing the top 10\% leaves too few tokens under the retention ratio of 10\%.}, and vary the removal ratio from 25\% to 75\%.

\smallskip
\noindent \textbf{Results. } As shown in~\cref{fig:naive_approach_no_add}, we observe that simply removing sink tokens (green line) yields competitive performance at removal ratio of 50\% and 75\% relative to the baseline (0\%), even under this reduced token budget. This indicates that sink tokens contribute little to fine-grained video understanding due to their low semantic content. On the other hand, replacing sink tokens with the next most attentive tokens (red line) substantially reduces hallucination, demonstrating that the benefit observed in Sec.~3.3 stems not from the absence of sink tokens per se, but from the introduction of more semantically salient tokens in their place. In summary, these results confirm that sink tokens are semantically inert, and that maximizing the proportion of truly salient tokens in the retained set is the key factor for fine-grained video understanding.

\begin{wrapfigure}{r}{0.39\linewidth}
  \centering
  \vspace{-26pt} 
  \includegraphics[width=0.99\linewidth]{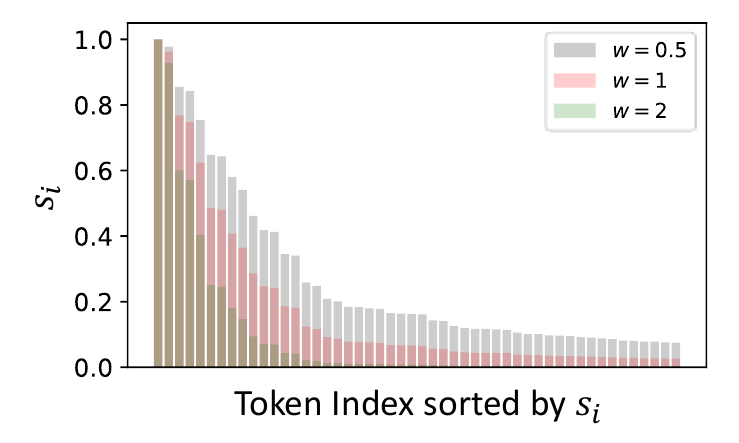}
  \vspace{-5.5ex}
  \caption{Performance degradation rate across varying sink token removal ratios.}
  \label{fig:w_value}
  \vspace{-21pt}
\end{wrapfigure}

\section{Distribution Sharpening via $w$}
\label{app_sec:dist_w}
In Eq.~(4) of the main paper, the hyperparameter $w$ controls the sharpness of the sink score distribution. Specifically, raising $\hat{s}_i$ to the power $w$ before min-max normalization amplifies the contrast between high-and low-sink scores when $w>1$, and diminishes it when $w<1$. As shown in~\cref{fig:w_value}, we visualize the normalized sink scores $s_i$ sorted in descending order for $w\in\{0.5, 1, 2\}$. At $w=0.5$, the distribution decays slowly, assigning relatively high scores to a large proportion of tokens and thus failing to clearly distinguish sink tokens from salient ones. As $w$ increases, the distribution becomes increasingly concentrated: a small subset of tokens retains scores near 1 while the majority are suppressed toward 0. This sharpening effect is desirable, as it allows the sink penalty in Eq.~(5) to be applied selectively to true sink tokens while minimizing unintended penalization of truly salient tokens. The quantitative analysis of $w$ is conducted in~\cref{app_sec:hyper}.

\section{Experiment}

\subsection{Additional Details of Datasets}
\label{app_sec:dataset_detail}
We provide further details, including evaluation metrics, for each benchmark used in the main paper.

\smallskip
\textbf{\textit{Fine-Grained Video Understanding Tasks}} 
\begin{itemize}[leftmargin=5mm]
\item \textbf{EventHallusion}~\cite{zhang2024eventhallusion} is a hallucination benchmark specifically designed for Video LLMs, comprising binary yes/no questions (Binary) that test susceptibility to language bias, and open-ended descriptive task (Desc.) that require fine-grained visual grounding. For both tasks, performance is reported as accuracy: Binary responses are evaluated directly against ground-truth labels, while Desc. responses are judged by GPT-5, which determines whether each generated description is correct or not.
\item \textbf{VideoComp}~\cite{kim2025videocomp} evaluates compositional reasoning, requiring models to identify subtle differences in action transitions and object states across frames. We report accuracy on two domains: ActivityNet (Act.) and YouCook2 (YC), where positive and negative sample pairs are distinguished by subtle differences in actions and object states.
\item \textbf{VCG-Bench}~\cite{maaz2024video} is an open-ended video question-answering benchmark evaluating five dimensions: Consistency (Consist.), Temporal Understanding (Temp.), Contextual Understanding (Context), Detail Orientation (Detail), and Correctness of Information (Correct). Model responses are scored by GPT-5 on a 0--5 scale, and we report the average score for each dimension.
\end{itemize}
\smallskip
\textbf{\textit{Multiple-Choice Question Answering}} 

We report accuracy for all these benchmarks.
\begin{itemize}[leftmargin=5mm]
\item \textbf{MVBench}~\cite{li2024mvbench} is a multiple-choice video understanding benchmark covering 20 temporal reasoning tasks.
\item \textbf{VideoMME}~\cite{fu2025video} is a multiple-choice video understanding benchmark spanning three temporal granularities: Short ($\leq$2 min), Medium (4--15 min), and Long (30--60 min).
\item \textbf{NextQA}~\cite{xiao2021next} is a multiple-choice benchmark focused on causal and temporal question answering with five answer candidates. 
\item \textbf{LongVideoBench}~\cite{wu2407longvideobench} evaluates long-form video-language understanding with 4 or 5 answer choices, requiring models to reason over extended temporal contexts.
\item \textbf{MLVU}~\cite{zhou2025mlvu} is a long video understanding benchmark covering diverse tasks including plot question answering, character identification, and anomaly detection.
\end{itemize}

\begin{wraptable}{r}{0.52\textwidth}
    \centering
\vspace{-8.0ex}
\caption{Performance of~\proposed{} applied to the concurrent method FlashVid.}
\resizebox{0.93\linewidth}{!}{
\begin{tabular}{l|c|c|c}
\toprule
\multirow{2}{*}{\textbf{Method}} & \multirow{2}{*}{\textbf{\begin{tabular}[c]{@{}c@{}}Retention\\ Ratio\end{tabular}}} & \multirow{2}{*}{\textbf{EventHallusion}} & \multirow{2}{*}{\textbf{VideoComp}} \\
                                 &                                                                                     &                                          &                                     \\ \midrule \midrule
Vanilla                          & 100\%                                                                               & 52.88                                    & 70.58                               \\ \midrule
FlashVid \scriptsize(ICLR'26)                        & \multirow{2}{*}{15\%}                                                               & 50.91                                    & 69.43                               \\
\cellcolor{gainsboro}FlashVid+\proposed{}             &                                                                                    & \cellcolor{gainsboro} \textbf{51.90}                                    & \cellcolor{gainsboro}\textbf{69.58}                               \\ \midrule
FlashVid  \scriptsize(ICLR'26)                       & \multirow{2}{*}{10\%}                                                               & 49.79                                    & 68.08                               \\
\cellcolor{gainsboro}FlashVid+\proposed{}             &                                                                                     & \cellcolor{gainsboro}\textbf{51.05}                                    & \cellcolor{gainsboro}\textbf{68.68}                               \\ \bottomrule
\end{tabular}
}
\label{app_tab:flashvid}
\vspace{-4ex}
\end{wraptable}

\subsection{Applying~\proposed{} to Concurrent Visual Token Pruning Method}
\label{app_sec:flash_vid}

To further demonstrate the adaptability of~\proposed{}, we apply it to FlashVid~\cite{fan2026flashvid}, a concurrent visual token pruning method that selects salient tokens per frame based on attention scores and diversity via the Max-Min Diversity Problem~\cite{alvar2025divprune}, followed by tree-based spatio-temporal merging. Specifically, we apply the STSP module by replacing the original attention scores with the sink-penalized scores $\tilde{A}_i^t$ described in Eq.~(5) of the main paper, and evaluate on EventHallusion and VideoComp at retention ratios of 10\% and 15\%. As shown in~\cref{app_tab:flashvid}, we observe that applying~\proposed{} to FlashVid consistently improves over the baseline, demonstrating that~\proposed{} is broadly applicable beyond the three primary baselines (VisionZip, FastVid, and Holitom) and can enhance concurrent state-of-the-art methods without any modification to their core architecture.

It is worth noting that the performance gains from~\proposed{} on FlashVid are smaller than those observed on VisionZip or FastVid in Tab.~1 of the main paper. We attribute this to FlashVid's tree-based merging, which reduces temporal redundancy and thus implicitly suppresses sink tokens—an effect analogous to the temporal pruning mechanism discussed in Sec.~3.3 of the main paper. Nevertheless,~\proposed{} still provides a meaningful and consistent benefit, confirming that explicit sink-aware pruning is complementary to the implicit effects of temporal merging strategies.

\subsection{Additional Implementation Details}
\label{app_sec:implementation}
For the hyperparameters of the baselines, we follow the prior baseline configurations. Regarding the hyperparameters $\mu_s$ and $\mu_t$ of the temporal+spatial pruning method Holitom, we perform a greedy search over $\mu_s\in \{0.01, 0.02, 0.03, 0.04\}$ and $\mu_t \in \{0.05,0.06,0.07,0.08 \}$ for each retention ratio. We set $\mu_s$ to 0.02 for FlashVid.

\begin{figure}[t]
    \centering
    \includegraphics[width=0.85\linewidth]{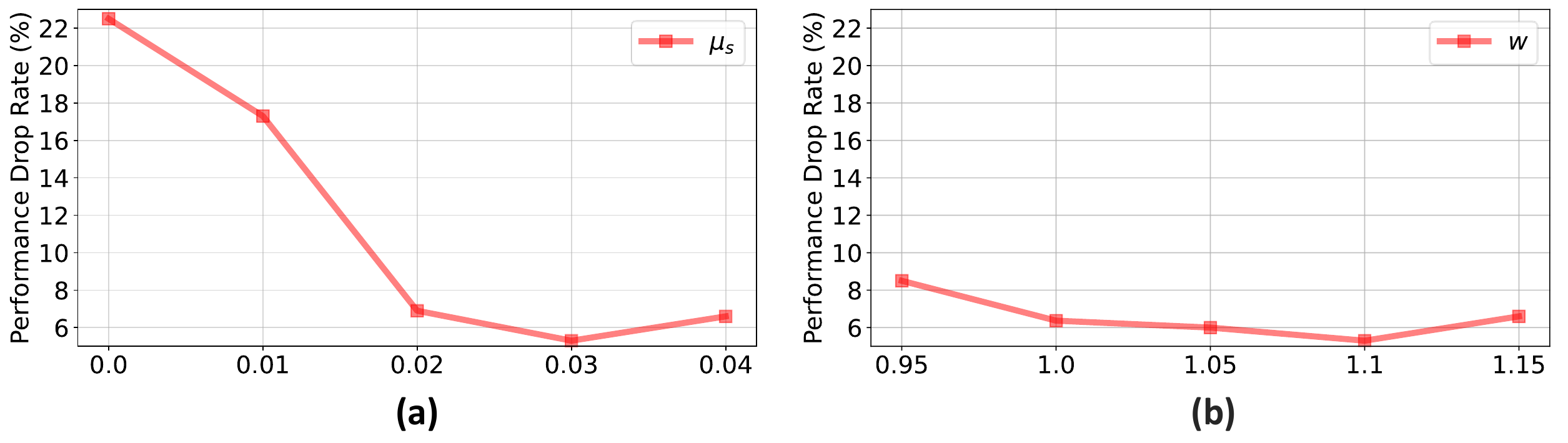}
    \vspace{-3.0ex}
    \caption{
    (a) Hyperparameter sensitivity of $\mu_s$ and (b) $w$ using VisionZip~\cite{yang2025visionzip} under retention ratio of 10\% on the EventHallusion~\cite{zhang2024eventhallusion} dataset.}   
    \label{app_fig:hyperparamter}
    \vspace{-3.5ex}
\end{figure}

\subsection{Hyperparameter Sensitivity}
\label{app_sec:hyper}
We analyze the sensitivity of three hyperparameters, $\mu_s$, $w$, and $\mu_t$, at a retention ratio of 10\% on the EventHallusion~\cite{zhang2024eventhallusion} dataset. Since $\mu_t$ is specific to the STTP module for temporal pruning, we use Holitom~\cite{shao2025holitom} for the sensitivity analysis of $\mu_t$, while VisionZip~\cite{yang2025visionzip} is used for $\mu_s$ and $w$.

\smallskip
\noindent \textbf{Sensitivity to $\mu_s$.} 
As shown in~\cref{app_fig:hyperparamter}(a), the performance drop rate decreases consistently as $\mu_s$ increases from 0.0\footnote{The variant with $\mu_s =0.0$ recovers the original VisionZip baseline without any sink suppression.} to 0.03, after which it slightly increases. We attribute this slight degradation beyond $\mu_s=0.03$ to over-penalization, where tokens near sink boundary that still carry useful semantic information are inadvertently suppressed. Accordingly, we identify $\mu_s=0.03$ as the optimal value, which effectively balances between suppressing sink tokens and preserving truly salient tokens.

\smallskip
\noindent \textbf{Sensitivity to $w$.}
As shown in~\cref{app_fig:hyperparamter}(b), the performance drop rate is relatively stable across $w\in [1.0, 1.15 ]$, with an optimum at $w=1.1$. As discussed in~\cref{app_sec:dist_w}, values of $w<1.0$ yield higher performance drop rates, as they diminish the contrast between high- and low-sink scores, making it difficult to distinguish sink tokens from salient ones. On the other hand, $w \geq 1.0$ benefits from the sharpening effect, concentrating high sink scores on true sink tokens while minimizing inadvertent penalization of salient tokens. Overall,~\proposed{} is more sensitive to $\mu_s$ than to $w$, indicating that $\mu_s$ is the more critical hyperparameter for effectively suppressing sink tokens.

\begin{wrapfigure}{r}{0.39\linewidth}
  \centering
  \vspace{-26pt} 
  \includegraphics[width=0.99\linewidth]{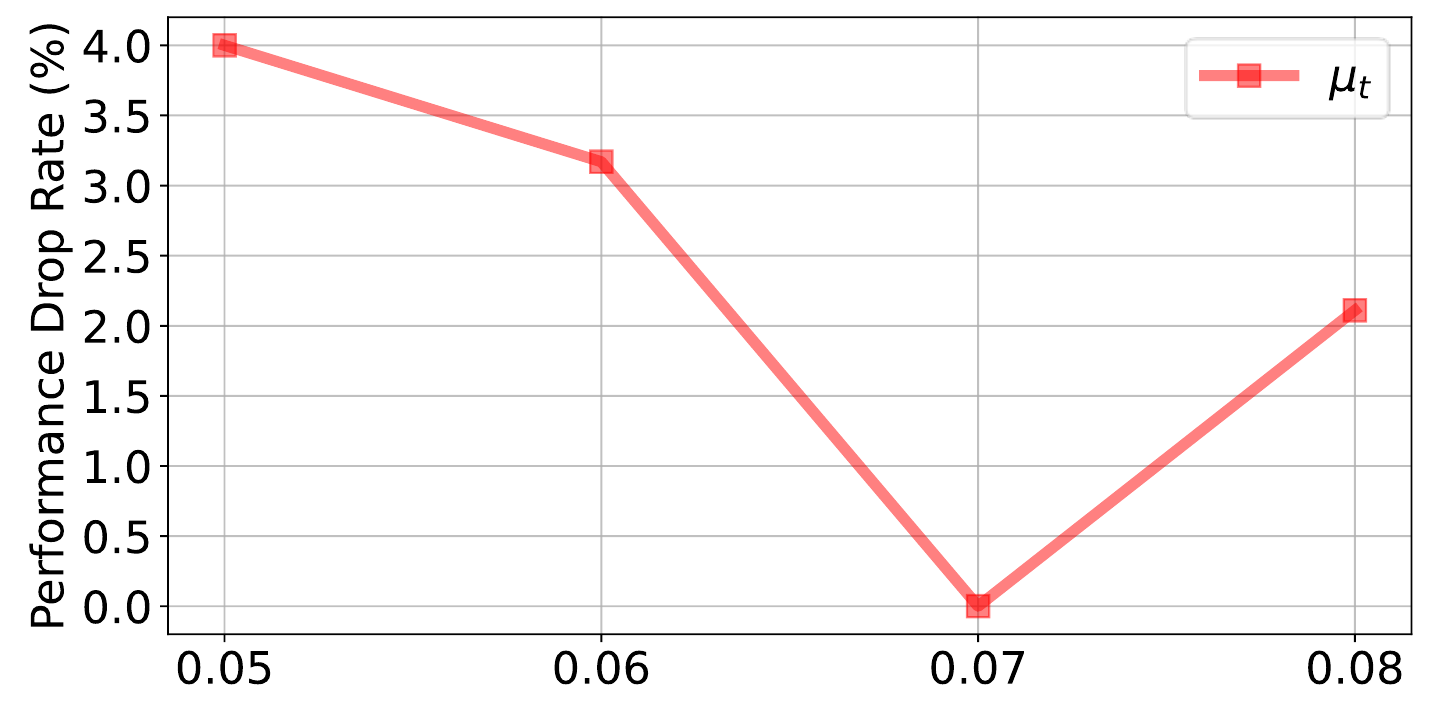}
  \vspace{-5.5ex}
  \caption{Hyperparameter sensitivity of $\mu_t$.}
  \label{fig:w_t_value}
  \vspace{-21pt}
\end{wrapfigure}

\smallskip
\noindent \textbf{Sensitivity to $\mu_t$.}
 As shown in~\cref{fig:w_t_value}, the performance drop rate decreases as $\mu_t$ increases from 0.05 to 0.07, beyond which it slightly increases. Similar to the over-penalization observed with large $\mu_s$, we attribute this degradation beyond $\mu_t=0.07$ to over-promotion of sink token pruning along the temporal axis. We therefore identify $\mu_t=0.07$ as the optimal value that effectively facilitates the temporal pruning of sink tokens while preserving truly salient tokens.

\section{Limitation and Future Work}
\label{app_sec:limitation}
In this work, we focus on suppressing sink tokens within attention-based pruning to retain salient tokens under a limited token budget. A potential limitation is that \proposed{} cannot be directly applied to feature-based pruning approaches~\cite{alvar2025divprune,huang2025prunevid}, as discussed in Sec.~5.4 of the main paper, since these approaches do not rely on attention scores. Nonetheless, as shown in Sec.~5.4 of the main paper,~\proposed{} applied to attention-based pruning still surpasses feature-based pruning, suggesting that explicitly identifying and suppressing problematic tokens is a more effective strategy than implicitly bypassing them. Building on this insight, we believe that feature-based pruning could be further improved by addressing tokens with \textit{high-norm features} that exhibit high pairwise similarity—tokens that are similarly treated as outliers in prior work~\cite{jiang2025vision,kim2025activation} and may collapse clustering algorithms such as DPC-KNN~\cite{du2016study}, playing an analogous role to sink tokens in the attention-based setting. As future work, we aim to extend the sink-aware pruning principle to feature-based pruning by identifying the analogous critical factor that hinders fine-grained video understanding in the feature space and explicitly addressing it during pruning.

\section{Broader Applicability of~\proposed{}}
\label{app_sec:general}

To further validate the generalization of~\proposed{}, we conduct additional experiments using a new dataset, Argus~\cite{rawal2025argus}, and a different backbone, Qwen2.5-VL~\cite{bai2025qwen25vltechnicalreport}.

\begin{wraptable}{r}{0.43\textwidth}
\centering
\vspace{-2.7ex}
\caption{Performance comparison on the Argus~\cite{rawal2025argus} dataset. $\downarrow$ indicates lower is better.}
\resizebox{0.93\linewidth}{!}{
\begin{tabular}{l|c|cc}
\toprule
\multirow{2}{*}{\textbf{Method}} & \multirow{2}{*}{\textbf{\begin{tabular}[c]{@{}c@{}}Retention\\ Ratio\end{tabular}}} & \multicolumn{2}{c}{\textbf{Argus}} \\
                                 &                                                                                     & Hallucination$\downarrow$      & Omission$\downarrow$     \\ \midrule \midrule
Vanilla                          & 100\%                                                                               & 55.08              & 73.30         \\ \midrule
FastVid                          & \multirow{4}{*}{10\%}                                                               & 51.54              & 79.01         \\
\cellcolor{gainsboro}FastVid+\proposed{}              &                                                                                     & \cellcolor{gainsboro}51.96              & \cellcolor{gainsboro}75.98         \\
VisionZip                        &                                                                                     & 56.19              & 80.57         \\
\cellcolor{gainsboro}VisionZip+\proposed{}            &                                                                                   & \cellcolor{gainsboro}52.42              & \cellcolor{gainsboro}77.29         \\ \bottomrule
\end{tabular}
}
\label{app_tab:additional_dataset}
\vspace{-4ex}
\end{wraptable}
\smallskip
\noindent \textbf{Experiments on Additional Dataset. } Beyond the benchmarks shown in the main paper, we evaluate on Argus~\cite{rawal2025argus}, a hallucination benchmark that measures two complementary aspects: Hallucination↓, which captures incorrectly generated content, and Omission↓, which captures relevant content missing from the generated response, where lower scores are better. As shown in~\cref{app_tab:additional_dataset}, applying~\proposed{} to VisionZip and FastVid reduces both scores. These results confirm that the benefits of explicitly suppressing sink tokens generalize across diverse hallucination benchmarks beyond those reported in the main paper.

\begin{wraptable}{r}{0.43\textwidth}
\centering
\vspace{-8.0ex}
\caption{Performance comparison using Qwen2.5-VL~\cite{bai2025qwen25vltechnicalreport} backbone.}
\resizebox{0.93\linewidth}{!}{
\begin{tabular}{l|c|ccc}
\toprule
\multicolumn{1}{c|}{\multirow{2}{*}{\textbf{Method}}} & \multirow{2}{*}{\textbf{\begin{tabular}[c]{@{}c@{}}Retention \\ Ratio\end{tabular}}} & \multicolumn{3}{c}{\textbf{EventHallusion}} \\ \cmidrule{3-5} 
\multicolumn{1}{c|}{}                                 &                                                                                      & Binary        & Desc.       & Overall       \\ \midrule \midrule
Vanilla                                               & 100\%                                                                                & 65.77         & 31.46       & 51.20         \\ \midrule
FastVid                                               & \multirow{4}{*}{10\%}                                                                & 61.61         & 25.17       & 46.13         \\
\cellcolor{gainsboro}FastVid+\proposed{}                                   &                                                                                      & \cellcolor{gainsboro}63.33         & \cellcolor{gainsboro}26.16       & \cellcolor{gainsboro}47.24         \\
VisionZip                                             &                                                                                      & 63.81         & 20.20       & 45.29         \\
\cellcolor{gainsboro}VisionZip+\proposed{}                                 &                                                                                      & \cellcolor{gainsboro}63.81         & \cellcolor{gainsboro}24.50       & \cellcolor{gainsboro}46.94         \\ \bottomrule
\end{tabular}
}
\label{app_tab:additional_backbone}
\vspace{-4ex}
\end{wraptable}

\smallskip
\noindent \textbf{Experiments on Additional Backbone.}
We evaluate VisionZip+\proposed{} and FastVid+\proposed{} using the Qwen2.5-VL~\cite{bai2025qwen25vltechnicalreport} backbone and compare against their respective baselines on EventHallusion at a 10\% retention ratio\footnote{Qwen2.5-VL inherently merges visual tokens prior to the LLM. We apply \proposed{} on top of these merged tokens and define the retention ratio relative to their count.}. As shown in~\cref{app_tab:additional_backbone},~\proposed{} consistently boosts the overall score compared to their respective baselines. However, we observe that the performance gains are smaller than those observed on 
LLaVA-OneVision~\cite{li2024llava} in the main paper.
We attribute this 
to Qwen2.5-VL's internal token merging, which inherently reduces 
temporal redundancy and thus implicitly suppresses the sink 
tokens—an effect analogous to the temporal pruning mechanism discussed in Sec.~3.3 of the main paper. Nevertheless, the consistent improvements confirm that sink tokens still persist even after merging and that \proposed{} effectively addresses this residual bottleneck, further demonstrating its applicability across different Video LLM backbones without any architectural modification.

\end{document}